\documentclass{article}

\usepackage[accepted]{mlsys2022}

\usepackage{color,xcolor}
\usepackage{epsfig}
\usepackage{graphicx}

\usepackage{adjustbox}
\usepackage{array}
\usepackage{booktabs}
\usepackage{colortbl}
\usepackage{wrapfig}
\usepackage{hhline}
\usepackage{multirow}
\usepackage{makecell}
\usepackage{printlen}
\usepackage{subcaption}
\usepackage{caption}

\usepackage{amsmath,amsfonts,amsthm,amssymb}
\usepackage{bm}
\usepackage{nicefrac}
\usepackage{microtype}
\usepackage{inconsolata}
\usepackage{pifont}

\usepackage{extramarks}
\usepackage{fancyhdr}
\usepackage{soul}
\usepackage{xspace}
\usepackage{titlesec}

\usepackage{hyperref}
\usepackage{url}

\usepackage{enumerate}
\usepackage{tikz}
\usepackage{xcolor}
\usepackage{lipsum}
\usepackage{listings}
\usepackage[firstpage]{draftwatermark}

\newcolumntype{L}[1]{>{\raggedright\let\newline\\\arraybackslash\hspace{0pt}}m{#1}}
\newcolumntype{C}[1]{>{\centering\let\newline\\\arraybackslash\hspace{0pt}}m{#1}}
\newcolumntype{R}[1]{>{\raggedleft\let\newline\\\arraybackslash\hspace{0pt}}m{#1}}

\newcommand{\sect}[1]{Section~\ref{#1}}

\newcommand{\eqn}[1]{Equation~\ref{#1}}
\newcommand{\fig}[1]{Figure~\ref{#1}}
\newcommand{\tbl}[1]{Table~\ref{#1}}
\newcommand{\algo}[1]{Algorithm~\ref{#1}}

\newcommand{\ignore}[1]{}

\makeatletter
\DeclareRobustCommand\onedot{\futurelet\@let@token\@onedot}
\def\@onedot{\ifx\@let@token.\else.\null\fi\xspace}

\def\eg{\emph{e.g}\onedot} 
\def\ie{\emph{i.e}\onedot} 
 
 \def\vs{\emph{vs}\onedot}

\makeatother

\def\SparseConv{SparseConv\xspace}

\def\sparseconvolution{sparse convolution\xspace}
\def\SparseConvolution{Sparse Convolution\xspace}
\def\sparsemapping{mapping\xspace}

\def\mappingpairs{maps\xspace}
\def\Mappingpair{Map\xspace}

\def\matmul{matrix multiplication\xspace}
\def\MatMul{Matrix Multiplication\xspace}
\def\bmm{batched matrix multiplication\xspace}

\def\mm{matmul\xspace}
\def\MM{MatMul\xspace}

\def\gathermmscatter{gather-matmul-scatter\xspace}

\def\spconv{SpConv\xspace}
\def\minkowskiengine{MinkowskiEngine\xspace}

\newcommand*\circled[1]{\tikz[baseline=(char.base)]{\node[shape=circle,fill,inner sep=1pt,scale=0.75] (char){\textcolor{white}{#1}};}}

\newcommand{\OffsetSet}[2]{\ensuremath{\bm{\Delta}^{#1}(#2)}}

\definecolor{mydarkblue}{rgb}{0,0.08,1}
\definecolor{mydarkgreen}{rgb}{0.02,0.6,0.02}
\definecolor{mydarkred}{rgb}{0.8,0.02,0.02}
\definecolor{mydarkorange}{rgb}{0.40,0.2,0.02}
\definecolor{mypurple}{RGB}{111,0,255}
\definecolor{myred}{rgb}{1.0,0.0,0.0}
\definecolor{mygold}{rgb}{0.75,0.6,0.12}
\definecolor{mydarkgray}{rgb}{0.66, 0.66, 0.66}
\definecolor{linkcolor}{RGB}{18,21,115}

\newcommand{\cmark}{\ding{51}}%
\newcommand{\xmark}{\ding{55}}%
\hypersetup{colorlinks,allcolors=linkcolor}

\SetWatermarkText{\hspace*{5in}\raisebox{10in}{
\includegraphics[width=0.75in]{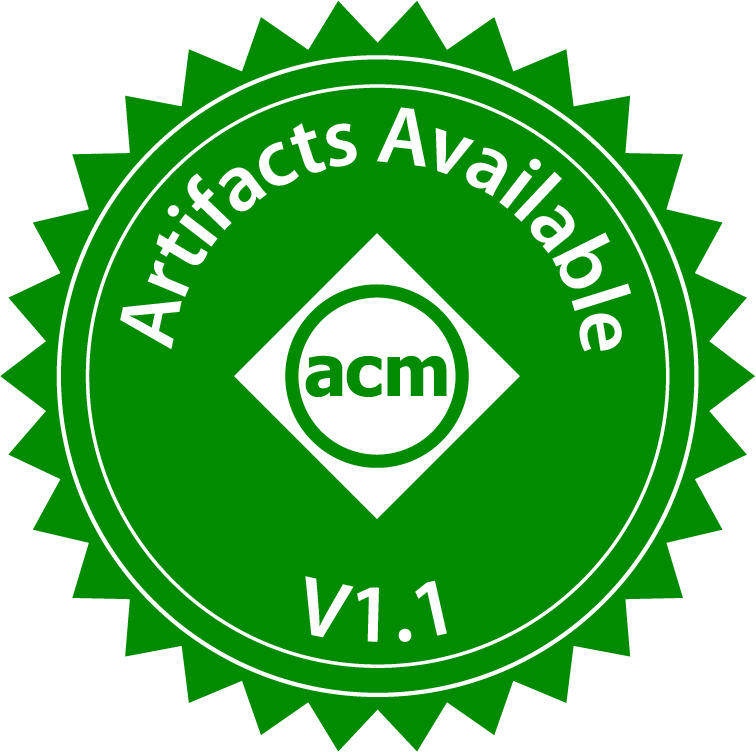}%
\includegraphics[width=0.75in]{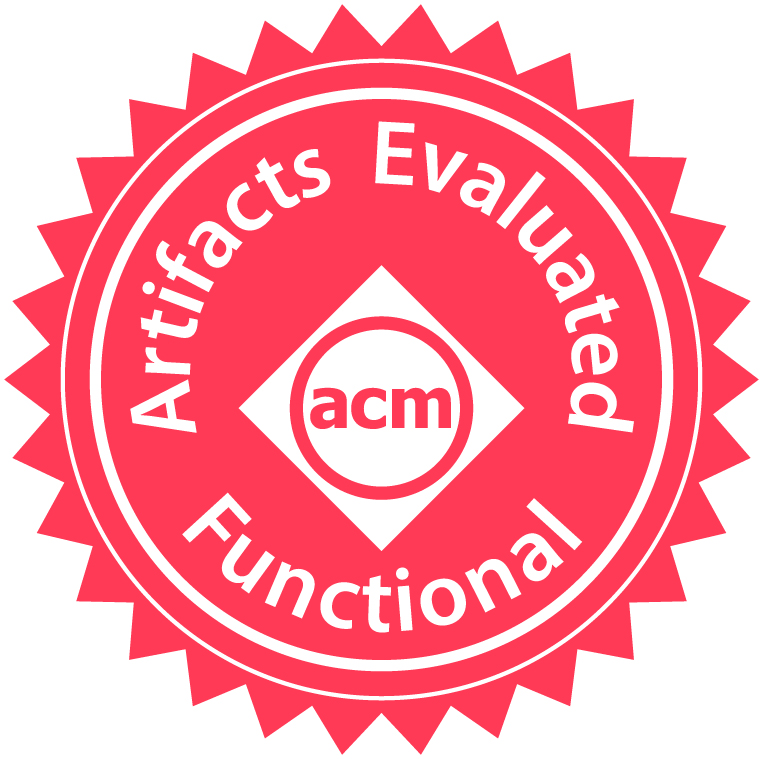}
}}
\SetWatermarkAngle{0}

\mlsystitlerunning{TorchSparse: Efficient Point Cloud Inference Engine}

\begin{document}

\twocolumn[
\mlsystitle{TorchSparse: Efficient Point Cloud Inference Engine}

\mlsyssetsymbol{equal}{*}

\begin{mlsysauthorlist}
\mlsysauthor{Haotian Tang}{equal,mit}
\mlsysauthor{Zhijian Liu}{equal,mit}
\mlsysauthor{Xiuyu Li}{equal,cornell}
\mlsysauthor{Yujun Lin}{mit}
\mlsysauthor{Song Han}{mit}
\end{mlsysauthorlist}

\mlsysaffiliation{mit}{Massachusetts Institute of Technology}
\mlsysaffiliation{cornell}{Cornell University}
\mlsyscorrespondingauthor{Song Han}{songhan@mit.edu}

\mlsyskeywords{Machine Learning, MLSys}

\begin{center}
    \url{https://torchsparse.mit.edu}
\end{center}

\vskip 0.3in

\begin{abstract}

Deep learning on point clouds has received increased attention thanks to its wide applications in AR/VR and autonomous driving. These applications require low latency and high accuracy to provide real-time user experience and ensure user safety. Unlike conventional dense workloads, the sparse and irregular nature of point clouds poses severe challenges to running sparse CNNs efficiently on the general-purpose hardware. Furthermore, existing sparse acceleration techniques for 2D images do not translate to 3D point clouds. In this paper, we introduce TorchSparse, a high-performance point cloud inference engine that accelerates the sparse convolution computation on GPUs. TorchSparse directly optimizes the two bottlenecks of sparse convolution: \textbf{irregular computation} and \textbf{data movement}. It applies \textit{adaptive matrix multiplication grouping} to trade computation for better regularity, achieving 1.4-1.5$\times$ speedup for matrix multiplication. It also optimizes the data movement by adopting \textit{vectorized}, \textit{quantized} and \textit{fused locality-aware memory access}, reducing the memory movement cost by 2.7$\times$. Evaluated on seven representative models across three benchmark datasets, TorchSparse achieves \textbf{1.6$\times$} and \textbf{1.5$\times$} measured end-to-end speedup over the state-of-the-art MinkowskiEngine and SpConv, respectively.

\end{abstract}
]

\printAffiliationsAndNotice{\mlsysEqualContribution}

\section{Introduction}

3D point cloud becomes increasingly accessible over the past few years thanks to the widely available 3D sensors, such as LiDAR scanners (on the self-driving vehicles and, more recently, even the mobile phones) and depth cameras (on the AR/VR headsets). Compared with 2D RGB images, 3D point clouds provide much more accurate spatial/depth information and are usually more robust to different lighting conditions. Therefore, 3D point cloud processing becomes the key component of many real-world AI applications: \eg, to understand the indoor scene layout for AR/VR, and to parse the driveable regions for autonomous driving.

\begin{figure}[t]
    \centering
    \includegraphics[width=\linewidth]{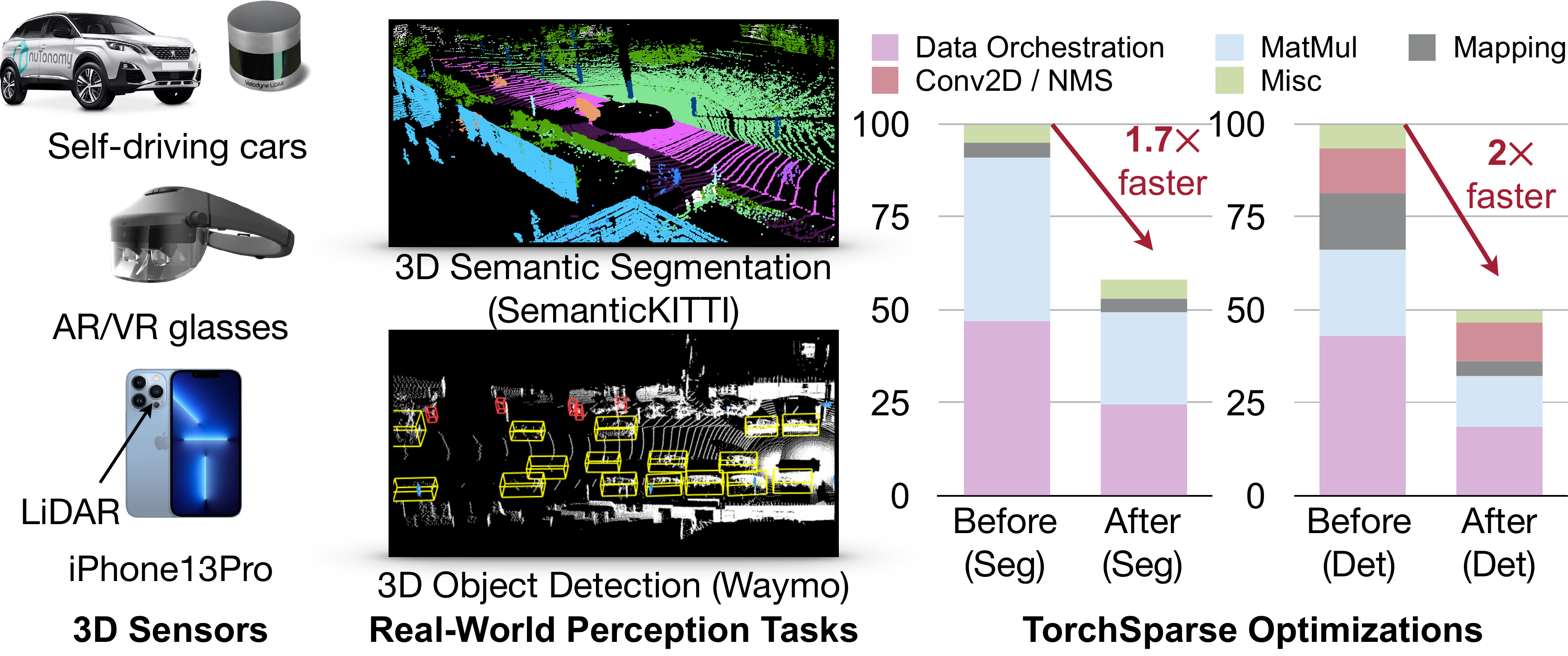}
    \caption{Widely available 3D sensors (left) have enabled more and more real-world AI applications to perceive the world using 3D point clouds (middle). However, processing 3D point cloud requires sparse computation that is not favored by general-purpose hardware. TorchSparse reduces the irregular computation and optimizes the data movement, achieving 1.7$\times$ to 2$\times$ measured speedup (right).}
    \label{fig:intro:teaser}
\end{figure}

A 3D point cloud is an \emph{unordered} set of 3D points. Unlike 2D image pixels, this data representation is highly sparse and irregular. Researchers have explored to rasterize the 3D data into dense volumetric representation~\cite{qi2016volumetric} or directly process it in the point cloud representation~\cite{qi2017pointnet++,li2018pointcnn}. However, both of them are not scalable to large indoor/outdoor scenes~\cite{liu2019pvcnn}. Alternatively, researchers have also investigated to flatten 3D point clouds into dense 2D representations using spherical projection and bird's-eye view (BEV) projection. However, their accuracy is much lower due to the physical dimension distortion and height information loss.

Recently, state-of-the-art 3D point cloud neural networks tend to rely largely or fully on sparse convolutions~\cite{graham20183d}, making it an important workload for machine learning system: all top 5 segmentation submissions on SemanticKITTI~\cite{behley2019semantickitti} are based on \SparseConv, 9 of top 10 submissions on nuScenes~\cite{caesar2020nuscenes} and top 2 winning solutions on Waymo~\cite{sun2020scalability} have exploited \SparseConv-based detectors~\cite{yin2021center,ge2021afdet}. Given the wide applicability and dominating performance of \SparseConv-based point cloud neural networks, it is crucial to provide efficient system support for \sparseconvolution on the general-purpose hardware.

Unlike conventional dense computation, sparse convolution is not supported by existing inference libraries (such as TensorRT and TVM), which is why most industrial solutions still prefer 2D projection-based models despite their lower accuracy. It is urgent to better support the sparse workload, which was not favored by modern high-parallelism hardware. On the one hand, the sparse nature of point clouds leads to irregular computation workloads: \ie, different kernel offsets might correspond to drastically different numbers of matched input/output pairs. Hence, existing sparse inference engines~\cite{yan2018second,choy20194d} usually execute the matrix multiplication for each kernel offset separately, which cannot fully utilize the parallelism of modern GPUs. On the other hand, neighboring points do not lie contiguously in the sparse point cloud representation. Explicitly gathering input features and scattering output results can be very expensive, taking up to 50\% of the total runtime. Due to the irregular computation workload and expensive data movement cost, SparseConv-based neural networks can hardly be run in real time: the latest sparse convolution library can only run MinkowskiNet at 8FPS on an NVIDIA GTX 1080Ti GPU, let alone other low-power edge devices.

In this paper, we introduce TorchSparse, a high-performance inference engine tailored for sparse point cloud computation. TorchSparse is optimized based upon two principles: (1) improving the computation regularity and (2) reducing the memory footprint. First, we propose the adaptive matrix multiplication grouping to batch the computation workloads from different kernel offsets together, trading \#FLOPs for regularity. Then, we adopt quantization and vectorized memory transactions to reduce memory movement. Finally, we gather and scatter features in locality-aware memory access order to maximize the data reduce. Evaluated on seven models across three datasets, TorchSparse achieves \textbf{1.6$\times$} and \textbf{1.5$\times$} speedup over state-of-the-art MinkowskiEngine and SpConv, paving the way for deploying 3D point cloud neural networks in real-world applications.
\section{Background}
\label{sect:background}

\begin{figure*}[t]
    \centering
    \includegraphics[width=\linewidth]{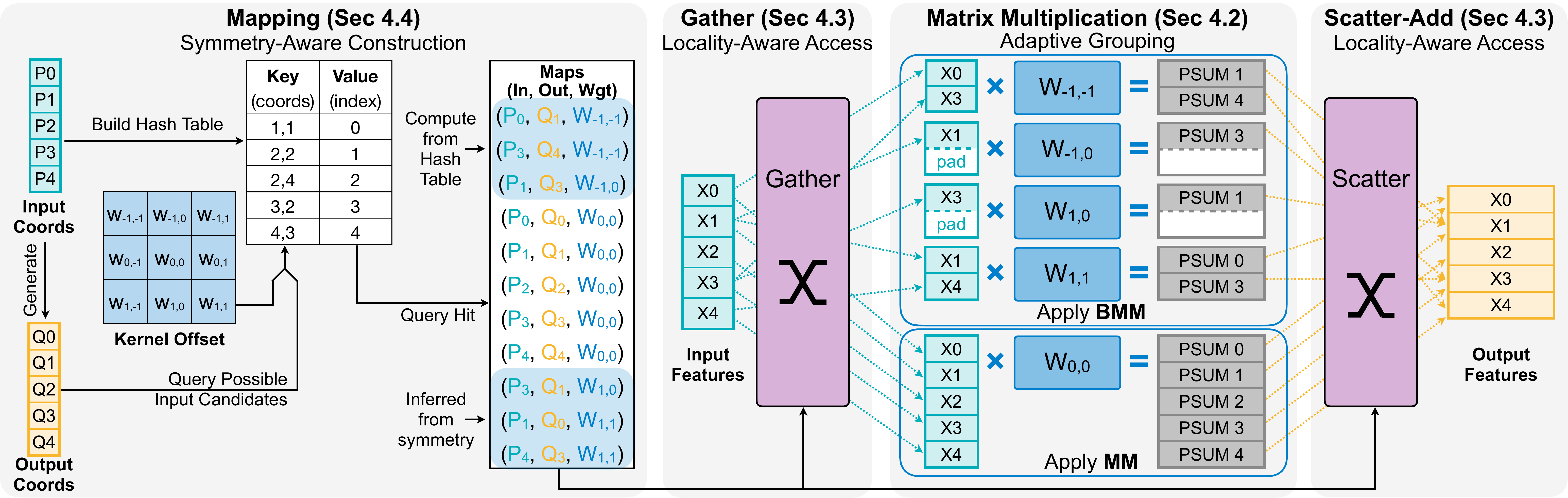}
    \caption{TorchSparse aims at accelerating \textit{\SparseConvolution}, which consists of four stages: mapping, gather, \mm and scatter-accumulate. Our optimization follows two principles: \protect\circled{1} improve the regularity of sparse workload \protect\circled{2} reduce the memory footprint. To achieve that, TorchSparse exploits adaptively batched \mm (Principle \protect\circled{1}, \sect{sect:method:matmul}); quantized, vectorized, locality-aware scatter/gather (Principle \protect\circled{2}, \sect{sect:method:scattergather}); and \sparsemapping kernel fusion (Principle \protect\circled{2}, \sect{sect:method:sparsemapping}).}
    \label{fig:background:dataflow}
\end{figure*}

\begin{figure}[t]
    \centering
    \includegraphics[width=\linewidth]{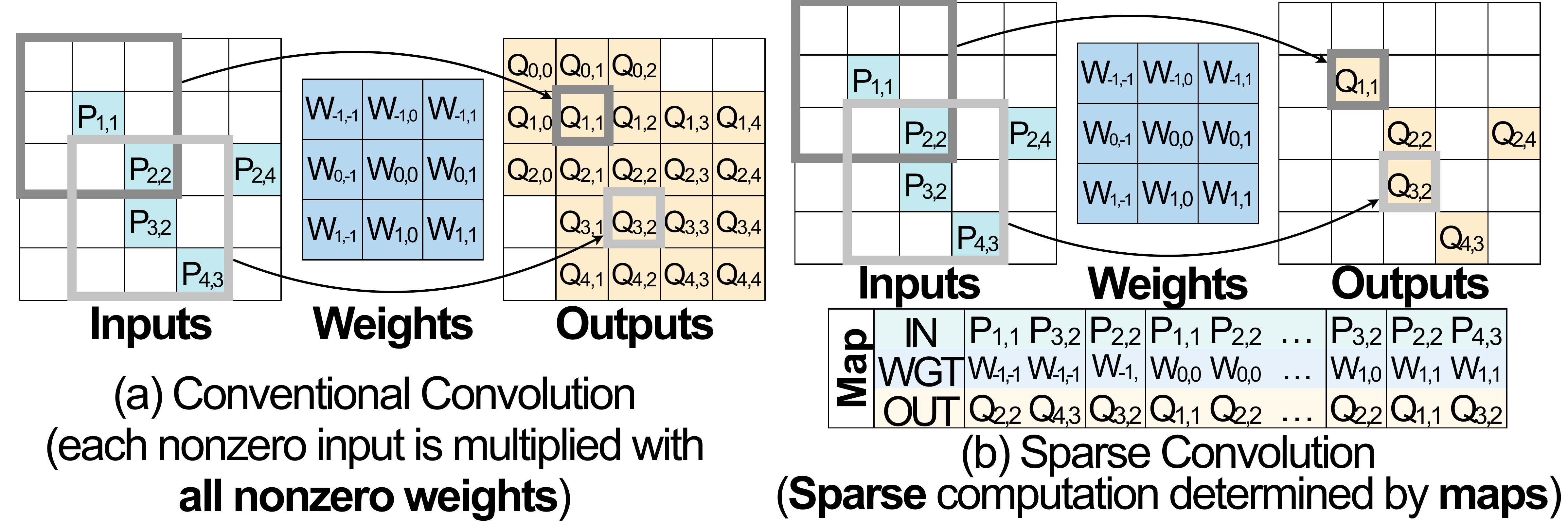}
    \caption{Sparse convolution (b) does \textit{not} multiply each nonzero input with all nonzero weights as conventional convolution (a) does.}
    \label{fig:background:conv}
\end{figure}

The point cloud can be formulated as an unordered set of points paired with features: $\{(\bm{p}_j, \bm{x}_{j})\}$, where $\bm{x}_{j} \in \mathbb{R}^{C}$ is a $C$-dimensional feature vector for point $\bm{p}_j \in \mathbb{Z}^D$ in a $D$-dimensional space. For a convolution of kernel size $K$, let $\bm{W} \in \mathbb{R}^{K^{D} \times C^{\text{in}} \times C^{\text{out}}}$ be its weights and \OffsetSet{D}{K} be its kernel offsets (\eg, $\OffsetSet{2}{5}$ = $\{-2, -1, 0, 1, 2\}^{2}$ and $\OffsetSet{3}{3}$ = $\{-1, 0, 1\}^{3}$). The weights $\bm{W}$ can be broken down into $K^D$ matrices of shape $C^{\text{in}}\times C^{\text{out}}$, denoted as $\bm{W}_{\bm{\delta}}$ for $\bm{\delta} \in \OffsetSet{D}{K}$.
With these notations, the convolution with stride $s$ can be represented as
\begin{equation} \label{eqn:conv}
    \bm{x}_{k}^{\text{out}} = \sum_{\bm{\delta} \in \OffsetSet{D}{K}}\sum_j 1[\bm{p}_j = s \bm{q}_k + \bm{\delta}]~(\bm{x}_{j}^{\text{in}} \cdot \bm{W}_{\bm{\delta}}),
\end{equation}
where $\bm{p}_j\in \bm{P}^{\text{in}}, \bm{q}_k \in \bm{P}^{\text{out}}$, and $1[\cdot]$ is the binary indicator.

For dense convolution (\fig{fig:background:conv}\textcolor{linkcolor}{a}), each nonzero input is multiplied with all nonzero weights, leading to rapidly growing nonzeros ($\bm{P}^{\text{in}}\subset\bm{P}^{\text{out}}$). On the other hand, the computation of sparse convolution (\fig{fig:background:conv}\textcolor{linkcolor}{b}) is determined by \textbf{\textit{maps}} $\mathcal{M} = \{(\bm{p}_j, \bm{q}_k, \bm{W}_{\bm{\delta}})\}$ in \eqn{eqn:conv} (also written as $\{(\bm{p}_j, \bm{q}_k, \bm{W}_n)\}$, where $n$ is the weight index) and keeps the sparsity pattern unchanged ($\bm{P}^{\text{in}}=\bm{P}^{\text{out}}$). It iterates over all maps and performs $\bm{x}_k^{\text{out}}=\bm{x}_k^{\text{out}}+\bm{x}_j^{\text{in}}\cdot\bm{W}_{\bm{\delta}}$.

\subsection{Mapping Operations}
\label{sect:background:mapping}

Mapping is a step to construct the input-output \emph{maps} $\mathcal{M} = \{(\bm{p}_{j}, \bm{q}_{k}, \bm{W}_{\bm{\delta}})\}$ for sparse convolution. Here, $j$ is the index of input point $\bm{p}$ in $\bm{P}^{\text{in}}$, $k$ is the index of output point $\bm{q}$ in $\bm{P}^{\text{out}}$, and $\bm{W}_{\bm{\delta}}$ is the weight matrix for kernel offset $\bm{\delta}$. Generating maps typically requires two steps: calculating the output coordinates $\bm{P}^{\text{out}}$, and searching maps $\mathcal{M}$. These operations only take coordinates as input.

\subsubsection{Output Coordinates Calculation}

When the convolution stride is 1, the output coordinates are exactly the same as the input coordinates, \ie, $\bm{P}^{\text{out}} = \bm{P}^{\text{in}}$.

When the convolution stride is larger than 1, the nonzero input coordinates will first be dilated for each kernel offset (\ie, $\bm{p}-\bm{\delta}$). After that, only these points on the strided grids within boundaries will become outputs $\bm{q}$, where $s\cdot\bm{q} = \bm{p}-\bm{\delta}$. 
Take the input coordinate $(3, 5)$ as an example (with stride of 2). For offset $\bm{\delta}=(1, 1)$, the output coordinate will be $\left((3, 5) - (1,1)\right)/2=(1,2)$, while for offset $\bm{\delta}=(0, 0)$, there is no valid output coordinate since $\left((3, 5) - (0,0)\right)$ is not a multiple of stride $s = 2$. We refer the readers to Appendix A for more details.

\subsubsection{\Mappingpair Search}
\label{sect:background:map_search}

As in \algo{alg:map_search}, map search requires iterating over all possible input coordinates for each output coordinate. A map is generated only when the input is nonzero.

\begin{algorithm}
\caption{\Mappingpair Search}\label{alg:map_search}
\textbf{Input:} input coordinates $\bm{P}^{\text{in}}$, output coordinates $\bm{P}^{\text{out}}$\\
\hspace*{2.8em} kernel size $K$, stride $s$\\
\textbf{Output:} maps $\mathcal{M}$
\begin{algorithmic}
\STATE $N \gets \OffsetSet{D}{K}.\texttt{size}$
\STATE $\mathcal{M}\gets \{\emptyset\}\times N$
\FOR {$k, \bm{q}_k$ \textbf{in} $\texttt{enumerate}(\bm{P}^{\text{out}})$}
  \STATE \small{\textcolor{blue}{\# Traverse the neighbors of an \textit{output} point.}}
  \FOR {$n, \bm{\delta}$ \textbf{in} $\texttt{enumerate}(\OffsetSet{D}{K})$}
  \STATE \small{\textcolor{blue}{\# Calculate \textit{input} coordinates.}}
  \STATE $\bm{r}\gets s\cdot\bm{q}_k+\bm{\delta}$
  \STATE \small{\textcolor{blue}{\# Add new map if input exists.}}
  \IF {$\bm{P}^{\text{in}}.\texttt{contain}(\bm{r})$}
    \STATE $j \gets \bm{P}^{\text{in}}.\texttt{getIndex}(\bm{r})$
    \STATE $\mathcal{M}[\bm{W}_{\bm{\delta}}]\gets \mathcal{M}[\bm{W}_{\bm{\delta}}]\cup\{(\bm{p}_j, \bm{q}_k, \bm{W}_{\bm{\delta}})\}$
  \ENDIF
  \ENDFOR
\ENDFOR
\end{algorithmic}
\end{algorithm}

To efficiently examine whether the possible input $\bm{q}_j + \bm{\delta}$ is nonzero, a common implementation is to record the coordinates of nonzero inputs with a hash table. The key-value pairs are (key=input coordinates, value=input index), \ie, $(\texttt{key}=\bm{p}_j, \texttt{value}=j)$. The hash function can simply be flattening the coordinate of each dimension into an integer.

\subsection{Data Orchestration and Matrix Multiplication}

After maps are generated, sparse convolution will multiply the input feature vector $\bm{x}_{j}^\text{in}$ with corresponding weight matrix $\bm{W}_{\bm{\delta}}$ and accumulate to the corresponding output feature vector  $\bm{x}_{k}^\text{out}$, following the map $\{(\bm{p}_{j}, \bm{q}_{k}, \bm{W}_{\bm{\delta}})\}$.

The utilization of matrix-vector multiplication is rather low on GPU. Therefore, most existing implementations follow the gather-matmul-scatter computation flow in \algo{alg:gather_mm_scatter}.
First, all input feature vectors associated with the same weight matrix are gathered and concatenated into a contiguous matrix. Then, matrix-matrix multiplication between feature matrix and weight matrix is conducted to obtain the partial sums. Finally, these partial sums are scattered and accumulated to the corresponding output feature vectors.

\begin{algorithm}
\caption{Gather-MatMul-Scatter}\label{alg:gather_mm_scatter}
\textbf{Input:} input features $\bm{X}^{\text{in}}$, weights $\bm{W}$, \mappingpairs $\mathcal{M}$\\
\textbf{Output:} output features $\bm{X}^{\text{out}}$
\begin{algorithmic}
\STATE $\bm{X}^{\text{out}} \gets \bm{0}$
\STATE \small{\textcolor{blue}{\# \textit{Separately} perform gather-\mm-scatter for each weight.}}
\FOR {$\bm{\delta}$ \textbf{in} $\OffsetSet{D}{K}$}
  \STATE $\bm{F} \gets \emptyset$
  \STATE \small{\textcolor{blue}{\# Gather features for $w_n$.}}
  \FOR {$m, (\bm{p}_j, \bm{q}_k, \bm{W}_{\bm{\delta}})$ \textbf{in} $\texttt{enumerate}(\mathcal{M}[\bm{W}_{\bm{\delta}}])$}
    \STATE $\bm{F}[m] \gets \bm{X}^\text{in}[j]$
  \ENDFOR
  \STATE \small{\textcolor{blue}{\# Matrix-matrix multiplication.}}
  \STATE $\bm{F} \gets \bm{F}\cdot\bm{W}_{\bm{\delta}}$
  \STATE \small{\textcolor{blue}{\# Scatter partial sums to $\bm{X}^{\text{out}}$.}}
  \FOR {$m, (\bm{p}_j, \bm{q}_k, \bm{W}_{\bm{\delta}})$ \textbf{in} $\texttt{enumerate}(\mathcal{M}[\bm{W}_{\bm{\delta}}])$}
    \STATE $\bm{X}^{\text{out}}[\bm{q}_k] \gets \bm{X}^{\text{out}}[\bm{q}_k] + \bm{F}[m]$
  \ENDFOR
\ENDFOR
\end{algorithmic}
\end{algorithm}

\subsection{Difference from Other Tasks}

\paragraph{\vs conventional convolution with sparsity.} The sparsity in conventional convolution comes from the ReLU activation function or weight pruning. Since there is no hard constraint on the output sparsity pattern, each nonzero input is multiplied with every nonzero weight, so the nonzeros will dilate during the inference, \ie, $\bm{P}^\text{in} \subset \bm{P}^\text{out}$ (see \fig{fig:background:conv}\textcolor{linkcolor}{a}).
The existing sparse computation libraries leverage such computation pattern by travelling all nonzero inputs with all nonzero weights to accelerate the conventional convolution.
On the contrary, sparse convolution requires $\bm{P}^\text{in} = \bm{P}^\text{out}$, and thus the relationship among inputs, weights and outputs requires to be explicitly searched with mapping operations, which makes it a hassle for previous sparse libraries.

\paragraph{\vs graph convolution.} In graph convolution, the relationship between inputs and outputs are provided in the adjacency matrix which stays constant across layers. Contrarily, sparse convolution has to search maps for \textbf{every} downsampling block. Furthermore, graph convolution shares the same weight matrix for different neighbors, \ie all $\bm{W}_{\bm{\delta}}$ are the same. Hence, graph convolution only needs either one gather or one scatter of features: 1) first gather input features associated with the same output vertex, and then multiply them with shared weights and reduce to the output feature vector; or 2) first multiply all input features with shared weights, and then scatter-accumulate the partial sums to the corresponding output feature vector. However, sparse convolution uses different weight matrices for different kernel offset $\bm{\delta}$ and thus needs both gather and scatter during the computation. Consequently, existing SpMV/SpDMM systems for graph convolution accleration~\cite{wang2019dgl,hu2020featgraph} are not applicable to \sparseconvolution.

\section{Analysis}
\label{sect:analysis}

\begin{figure}
    \centering
    \includegraphics[width=0.9\linewidth]{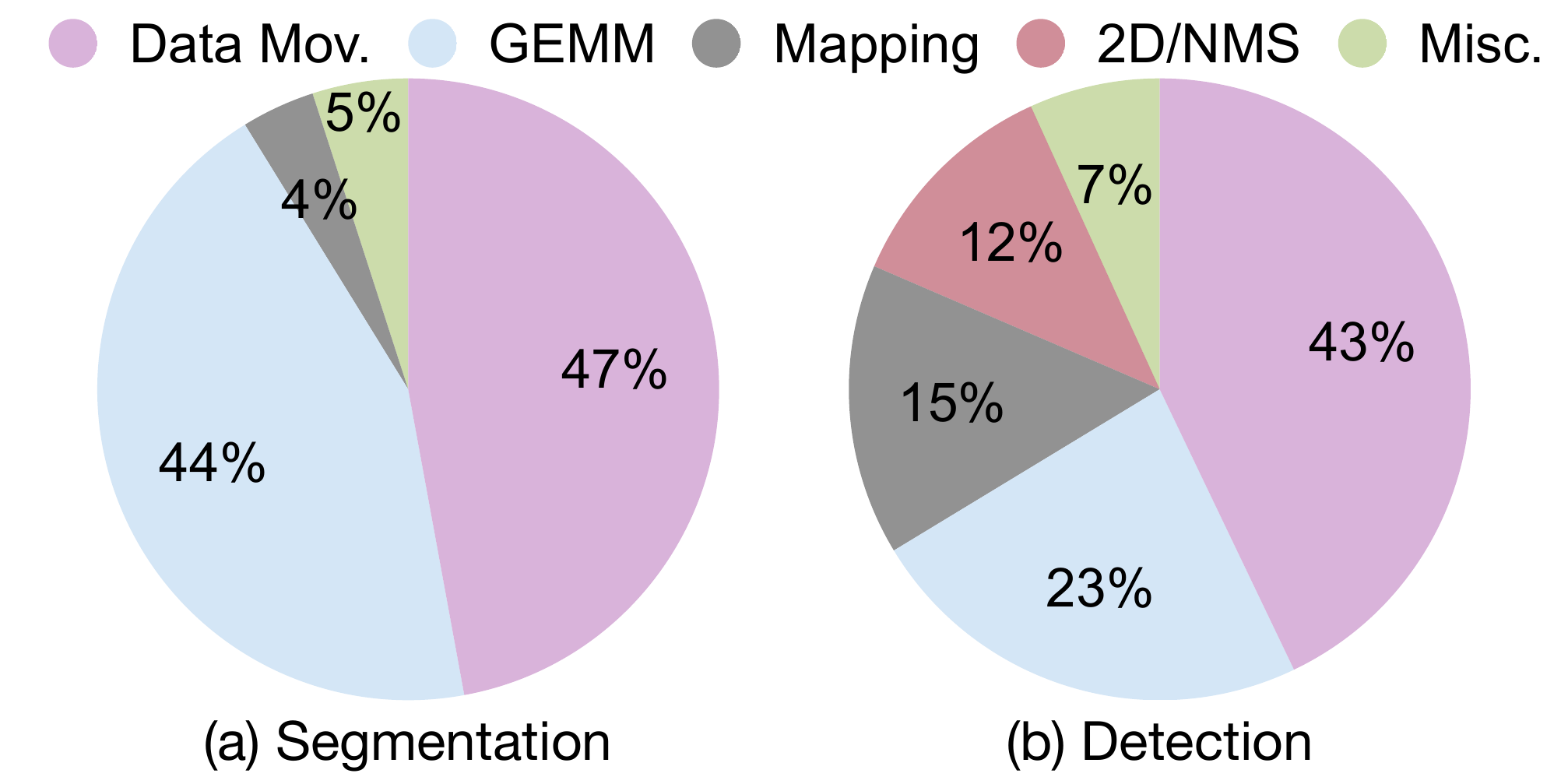}
    \caption{Data movement and GEMM constitute a significant proportion of the runtime of sparse CNNs.}
    \label{fig:analysis:breakdown}
\end{figure}

We systematically profile the runtime of different components in two representative sparse CNNs: one for segmentation (\fig{fig:analysis:breakdown}\textcolor{linkcolor}{a}) and one for detection (\fig{fig:analysis:breakdown}\textcolor{linkcolor}{b}). Based on observations in \fig{fig:analysis:breakdown}, we summarize two principles for sparse convolution optimization which lays the foundation for our system design in \sect{sect:method}.

\paragraph{Principle I. Improve Regularity in Computation}

Matrix multiplication is the core computation in sparse convolution and takes up a large proportion of total execution time (20\%-50\%). \algo{alg:gather_mm_scatter} decouples the matrix multiplication computation from data movement so that we can use well-optimized libraries (such as cuDNN) to calculate $\bm{X}^{\text{out}}\leftarrow \bm{X}^{\text{in}}\cdot\bm{W}_{\bm{\delta}}$. However, the computation workloads are very \textit{non-uniform} due to the irregular nature of point clouds (detailed in \fig{fig:ablation:mmsize}, where map sizes for different weights can differ by an order of magnitude, and most map sizes are \textit{small}). As a result, the matrix multiplication in MinkUNet (0.5$\times$ width) runs at 8.1 TFLOP/s on RTX 2080 Ti with FP16 quantization, achieving only 30\% device utilization. Therefore, \textit{improving the regularity} of matrix multiplication will potentially be helpful: we boost the utilization to 44.2\% after optimization (detailed in \tbl{tab:ablation:mm}).

\paragraph{Principle II. Reduce Memory Footprint}

Data movement is the largest bottleneck in sparse CNNs, which takes up 40\%-50\% of total runtime on average. This is because scatter-gather operations are bottlenecked by GPU memory bandwidth (\textit{limited}) rather than computation resources (\textit{abundant}). Worse still, the dataflow in \algo{alg:gather_mm_scatter} completely separates scatter-gather operations for different kernel offsets. This further ruins the possibility of any reuse in the data movement, which will be detailed in \fig{fig:method:locality}. It is also noteworthy that the large mapping latency in the CenterPoint detector (\fig{fig:analysis:breakdown}\textcolor{linkcolor}{b}) also stems from memory overhead: hashmap construction and  output coordinate calculation both require multiple DRAM accesses. Thus, \textit{reducing memory footprint} is at the heart of data movement and \sparsemapping optimization. 
\section{System Design and Optimization}
\label{sect:method}

\begin{figure*}[t]
    \centering
    \includegraphics[width=0.95\linewidth]{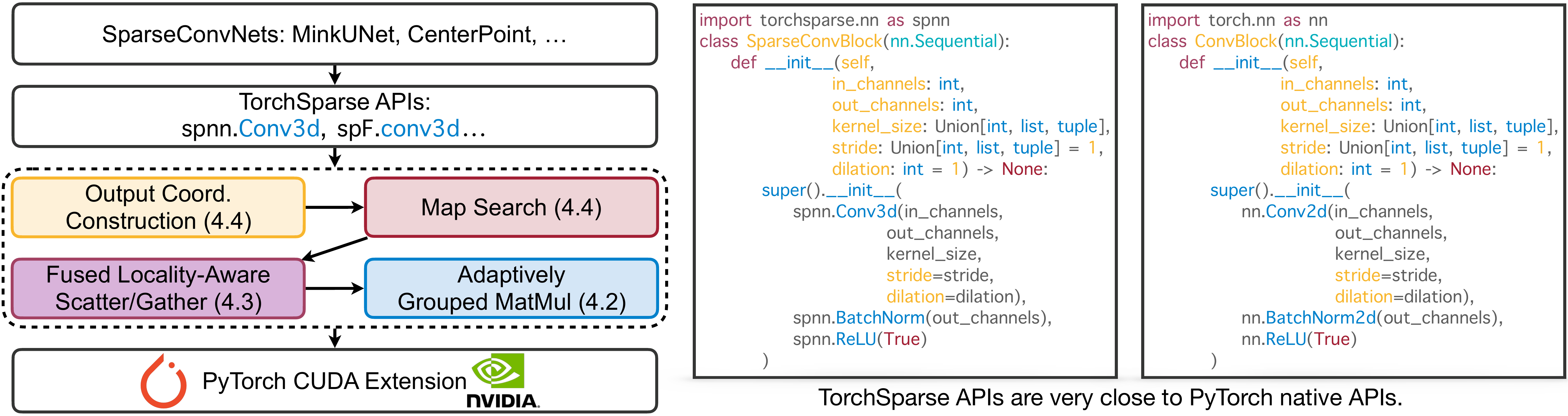}
    \caption{System overview for TorchSparse: our TorchSparse provides handy Python APIs similar to PyTorch and applies low-level optimizations to data movement, \matmul and \sparsemapping operations in sparse convolution.}
    \label{fig:method:api}
\end{figure*}

This section unfolds our TorchSparse as follows: \sect{sect:method:overview} provides an overview and API design for TorchSparse, \sect{sect:method:matmul} introduces improvements on matrix multiplication operations, \sect{sect:method:scattergather} elaborates the optimizations for data movement operations (scatter/gather), and \sect{sect:method:sparsemapping} analyzes the opportunities to speed up mapping operations.

\subsection{System Overview}
\label{sect:method:overview}

\fig{fig:method:api} provides an overview of our TorchSparse. At the top level, users define their sparse CNNs using TorchSparse APIs, which have minimal differences with native PyTorch APIs. Also, TorchSparse does not require users to add additional fields such as \texttt{indice\_key} and \texttt{spatial\_shape} in SpConv~\cite{yan2018second}, and \texttt{coordinate\_manager} in MinkowskiEngine~\cite{choy20194d} when defining modules and tensors. TorchSparse converts the high-level modules to primitive operations: \eg, \texttt{Conv3d} is decomposed to output construction, mapping operations and \gathermmscatter. For each part, Python APIs interact with backend CUDA implementations via \texttt{pybind}. Note that TorchSparse also provides support for CPU inference and multi-GPU training, but this paper will focus on the GPU inference.

\subsection{Matrix Multiplication Optimization}
\label{sect:method:matmul}

\begin{figure}[t]
\centering

\begin{subfigure}[t]{0.49\linewidth}
    \centering
    \includegraphics[width=\linewidth]{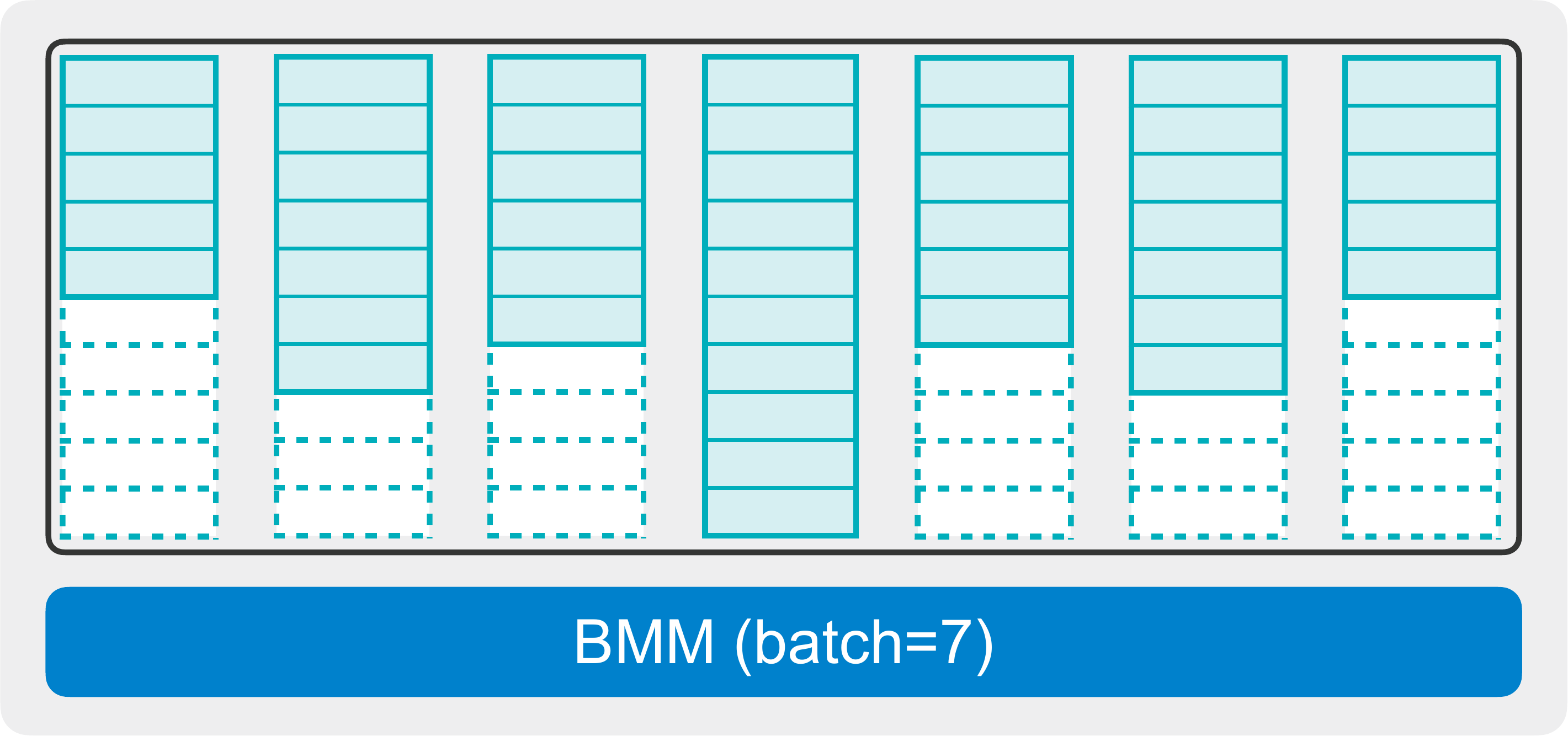}
    \caption{Dense Computation}
    \label{fig:method:grouping:dense}
\end{subfigure}
\hfill
\begin{subfigure}[t]{0.49\linewidth}
    \centering
    \includegraphics[width=\linewidth]{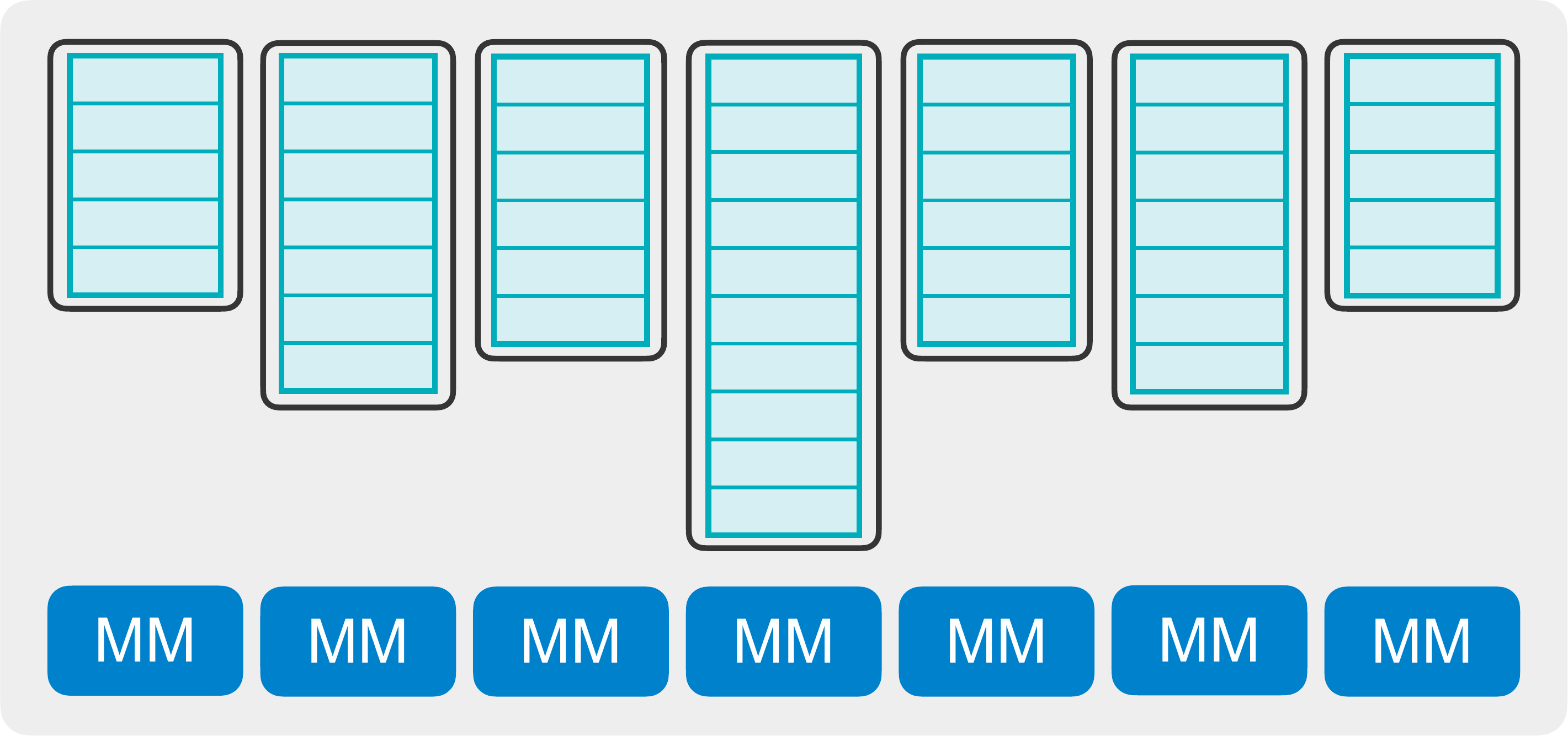}
    \caption{Separate MatMul}
    \label{fig:method:grouping:separate}
\end{subfigure}
\hfill
\begin{subfigure}[t]{\linewidth}
    \centering
    \includegraphics[width=\linewidth]{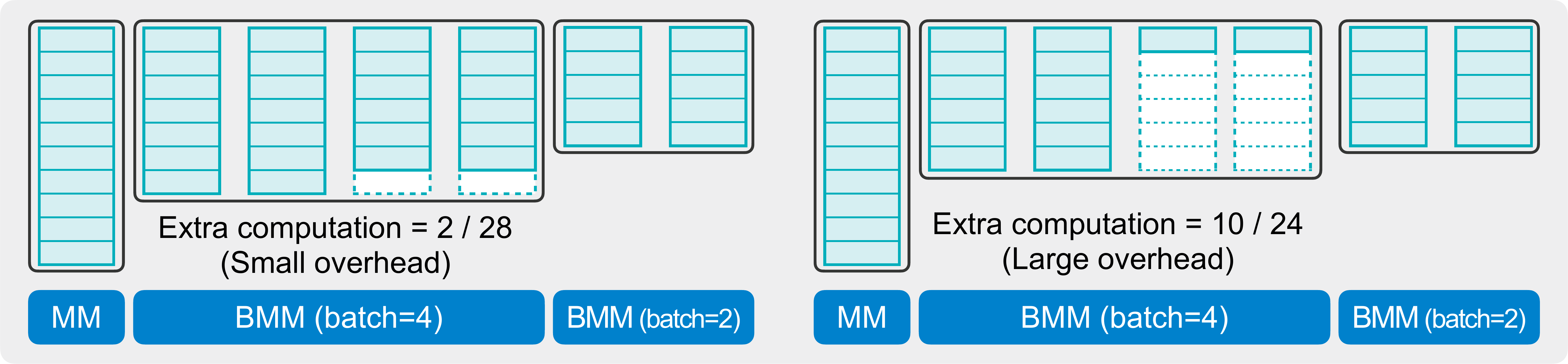}
    \caption{Fixed Grouping}
    \label{fig:method:grouping:fixed}
\end{subfigure}
\hfill
\begin{subfigure}[t]{\linewidth}
    \centering
    \includegraphics[width=\linewidth]{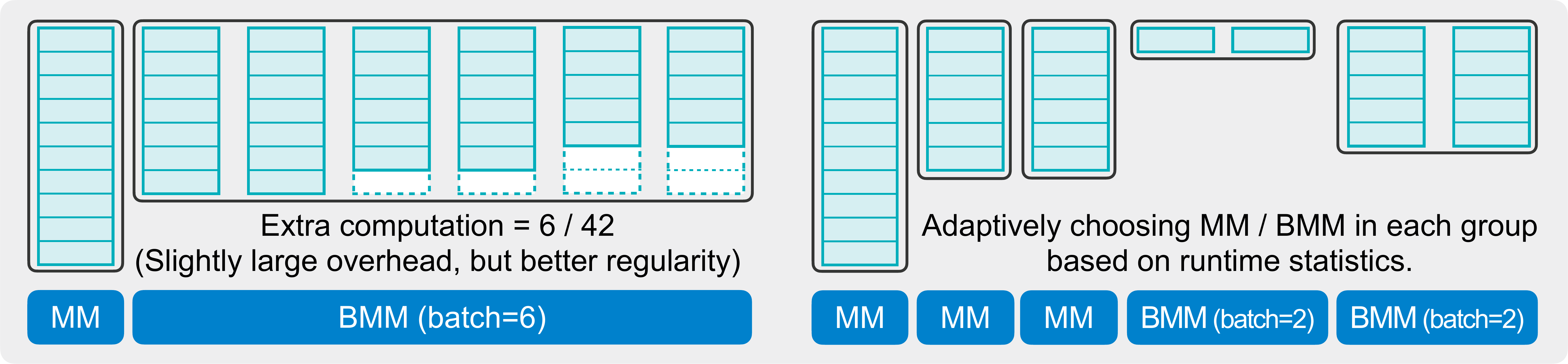}
    \caption{Adaptive Grouping}
    \label{fig:method:grouping:adaptive}
\end{subfigure}
\caption{Different matrix multiplication grouping strategies: (a) dense computation suffers from large FLOPs overhead; (b) separate matrix multiplication suffers from low device utilization and excessive kernel calls; (c) fixed grouping trades FLOPs for regularity; (d) adaptive grouping  searches for the best balance point.}
\label{fig:method:grouping}
\end{figure}

\begin{figure}
    \centering
    \includegraphics[width=0.85\linewidth]{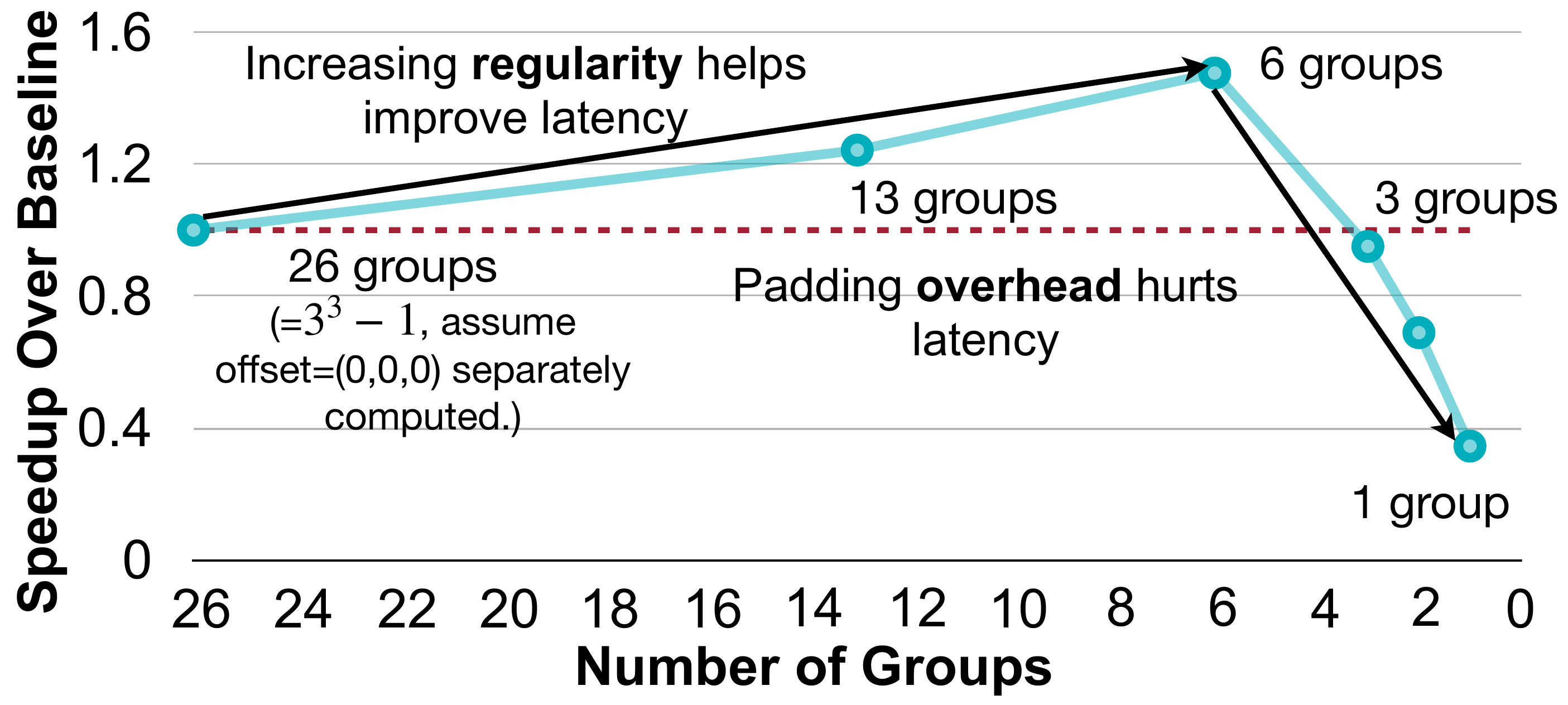}
    \caption{Trading FLOPs for computation regularity via batched \matmul brings 1.5$\times$ speedup.}
    \label{fig:method:mm_utilization}
\end{figure}

Matrix multiplication is the core computation in sparse convolution. Due to the irregularity of point clouds, existing implementations rely on cuDNN to perform many small matrix multiplications on different weights (\fig{fig:method:grouping:separate}), which usually do not saturate the utilization of GPUs. In order to increase the utilization, we propose to trade computation for regularity (Principle I) by grouping matrix multiplication for different weights together. We find it helpful to introduce redundant computation but group more computation in a single kernel. In \fig{fig:ablation:mmsize}, we collect the real workload for MinkUNet~\cite{choy20194d} on SemanticKITTI~\cite{behley2019semantickitti} and analyze the efficiency of matrix multiplication in the first sparse convolution layer with respect to the group size. It turns out that \bmm can be significantly faster than sequentially performing the computation along the batch dimension, thanks to the better regularity. This motivates us to explore the opportunity of \textit{grouping} in the matrix multiplication computation.

\subsubsection{Symmetric Grouping}

With sparse workloads, the map sizes for different weights within one sparse convolution layer are usually \textit{different}. Fortunately, for sparse convolutions with odd kernel size and stride of 1, the maps corresponding to kernel offset $(a, b, c)$ will always have the same size as the maps corresponding to the \textit{symmetric} kernel offset $(-a, -b, -c)$. For a map entry $(\bm{p}_j, \bm{q}_k, \bm{W}_{a, b, c})$, we have $\bm{q}_k = \bm{p}_j + (a, b, c)$. Then, $\bm{p}_j = \bm{q}_k + (-a, -b, -c)$, which implies that $(\bm{q}_k, \bm{p}_j, \bm{W}_{-a,-b,-c})$ is also a valid map entry. As such, we can establish an one-to-one correspondence between maps for weights $\pm(a, b, c)$. Therefore, we are able to group the workload for symmetric kernel offsets together and naturally have a batch size of 2. Note that the workload corresponding to the kernel offset $(0, 0, 0)$ is processed separately since it does not require any explicit data movement. From \fig{fig:method:mm_utilization}, the symmetric grouping (13 groups) can already be up to 1.2$\times$ faster than the separate matrix multiplication.

\subsubsection{Fixed Grouping}

Though symmetric grouping works well for sparse convolutions with the stride of 1, it falls short in generalizing to downsampling layers. Also, it cannot push the batch size to $>$ 2, which means that we still have a large gap towards the best GPU utilization in \fig{fig:method:mm_utilization}. Nevertheless, we find that clear pattern exists in the map size statistics (\fig{fig:ablation:mmsize}): for submanifold layers, the maps corresponding to $\bm{W}_0$ to $\bm{W}_3$ tend to have similar sizes and the rest of the weights other than the middle one have similar sizes; for downsampling layers, the maps for all offsets have similar sizes. Consequently, we can batch the computation into three groups accordingly. Within each group, we pad all features to the maximum size (\fig{fig:method:grouping:fixed}). Fixed grouping generally works well when all features within the same group have similar sizes (\fig{fig:method:grouping:fixed} left), and this usually happens in downsampling layers. For submanifold layers (\fig{fig:method:grouping:fixed} right), the padding overhead can sometimes be large despite the better regularity, resulting in wasted computation. 

\subsubsection{Adaptive Grouping}

The major drawback of fixed grouping is that it does \textit{not} adapt to individual samples. This can be problematic since workload size distributions can vary greatly across different datasets (\fig{fig:ablation:mmsize}). It is also very labor-intensive to design different grouping strategies for different \textit{layers}, different \textit{networks} on a diverse set of \textit{datasets} and \textit{hardware}. To this end, we design an adaptive grouping algorithm (\fig{fig:method:grouping:adaptive}) that \textit{automatically} determines the \textit{input-adaptive} grouping strategy for a given layer on arbitrary workload.

The adaptive grouping algorithm builds upon two auto-tuned parameters $\epsilon$ and $S$, where $\epsilon$ indicates our tolerance of redundant computation, and $S$ is the workload threshold. Given $\epsilon$, we scan over sizes of all maps in the current workload for $\bm{W}_0$ to $\bm{W}_{K-1}$ (where $K$ is the kernel volume) and dynamically maintain two pointers indicating the start and end of the current group. We initiate a new group whenever the redundant computation ratio ($1- \frac{\text{Theoretical FLOPs}}{\text{Actual FLOPs}}$) exceeds $\epsilon$. Then, given $S$, we inspect the maximum workload size within each group. Each group performs \texttt{bmm} if the workload size is smaller than $S$ and performs \texttt{mm} otherwise. This is because \texttt{bmm} can improve device utilization for small workloads but has little benefit for large workloads. We refer the readers to Appendix B for more details of this algorithm. Note that even if $\epsilon$ and $S$ are fixed, the generated strategy itself is still \textit{input-adaptive}. Since different input point clouds have different map sizes, even the same $\epsilon$ can potentially generate different group partition strategies for different samples. The $(\epsilon, S)$ parameter space is simple but diverse enough to cover dense computation ($\epsilon=1$; $S=+\infty$), separate computation ($S=0$) as well as symmetric grouping ($\epsilon=0; S=+\infty$) as its special cases. 

For a given sparse CNN, we determine $(\epsilon, S)$ for each layer on the target dataset and hardware platform via exhaustive grid search on a small subset (usually 100 samples) of the training set. We formalize this process in Appendix B. The search is inference-only. It explores a space of around 1,000 configurations and requires less than 10 minutes of search time on a desktop GPU. The strategy derived on the small subset can be directly applied and does not require any parameter optimization during the inference time.

\subsection{Data Movement Optimization}
\label{sect:method:scattergather}

\begin{figure}[t]
    \centering
    \includegraphics[width=0.9\linewidth]{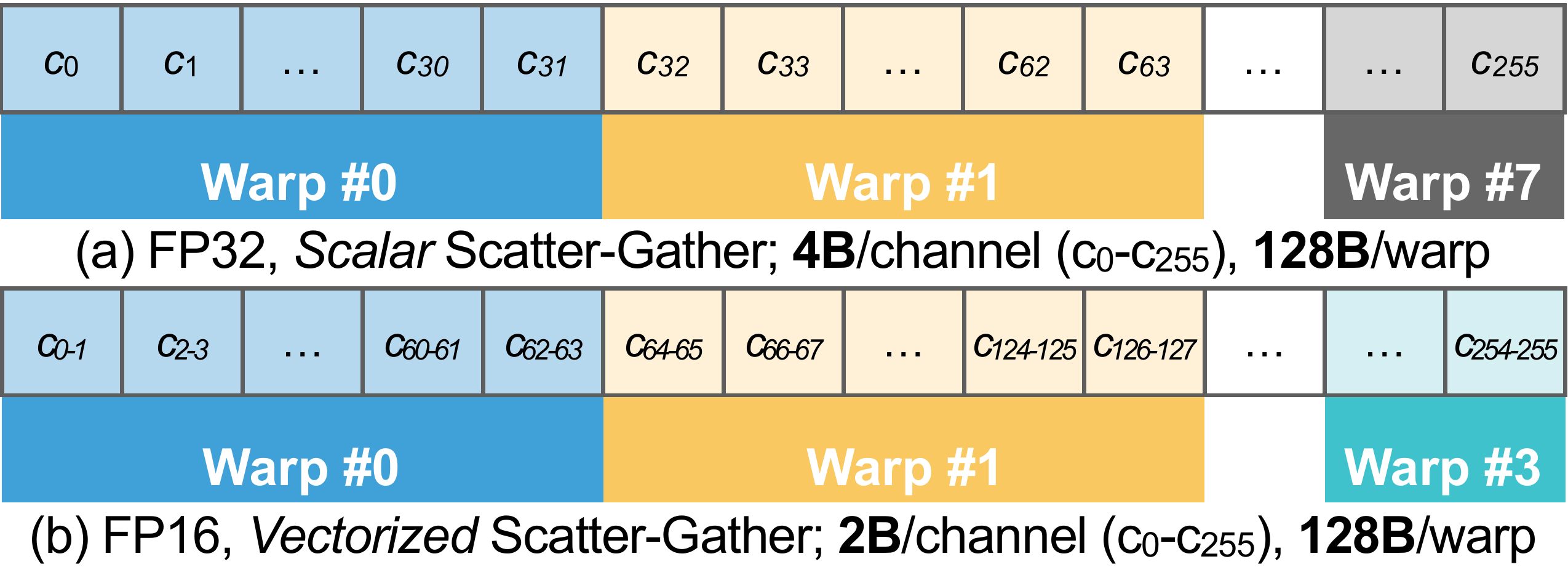}
    \caption{TorchSparse applies \textit{vectorized} and \textit{quantized} scatter-gather to greatly reduce the data movement latency.}
    \label{fig:method:fp16}
\end{figure}
\begin{figure}[t]
    \centering
    \includegraphics[width=0.9\linewidth]{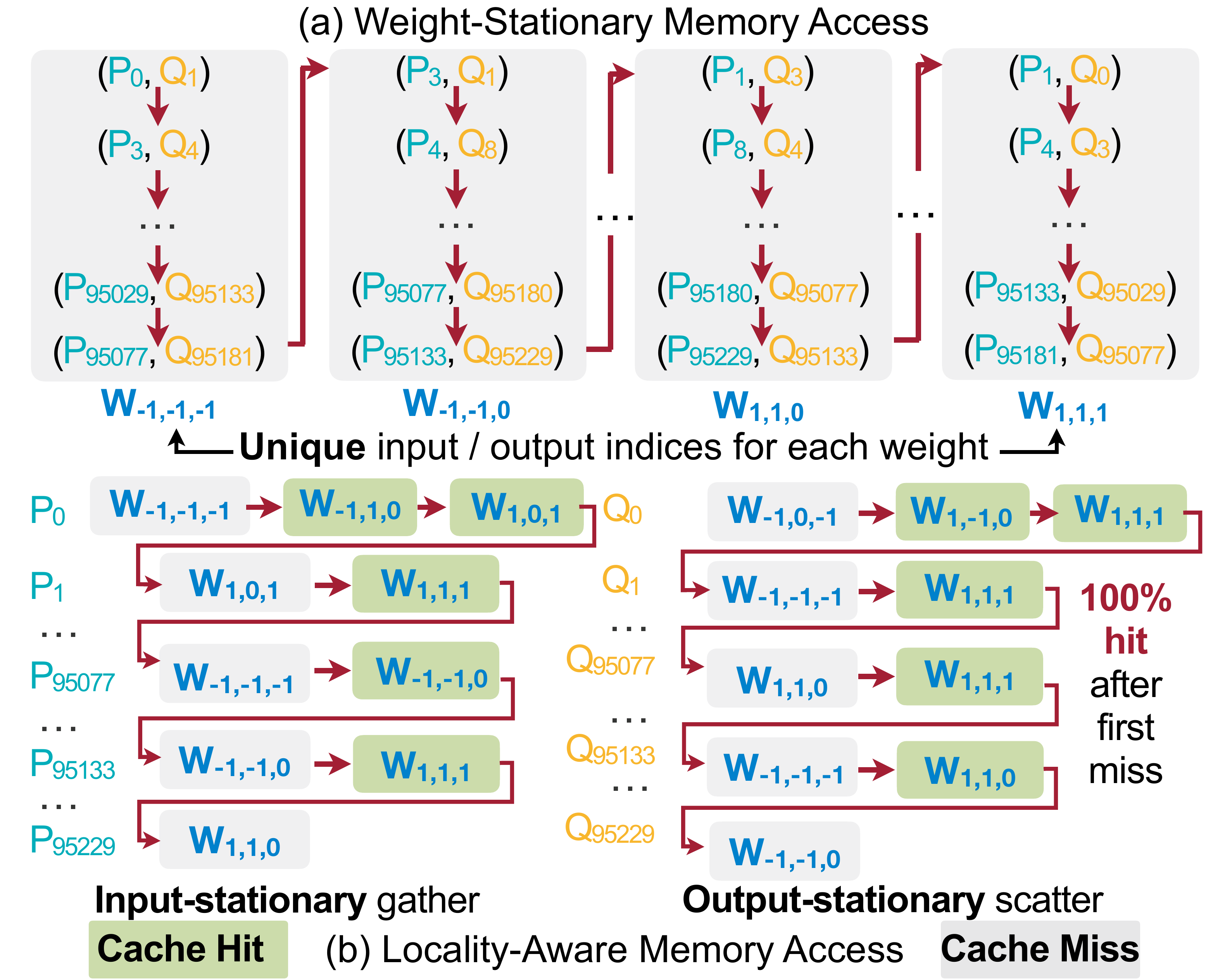}
    \caption{TorchSparse proposes cache-friendly \textit{locality-aware} and memory access pattern. In contrary, baseline implementation (a) cannot exploit cache reuse due to uniqueness in input/output indices for each weight.}
    \label{fig:method:locality}
\end{figure}

From \sect{sect:analysis}, data movement usually takes up 40-50\% of the total runtime. Thus, optimizing data movement will be of high priority as well (Principle II). Intuitively, it is most effective to reduce data movement cost by reducing the total amount of DRAM access and exploiting the data reuse.

\subsubsection{Quantized and Vectorized Memory Access}

FP16 quantization brings 2$\times$ theoretical DRAM access saving compared with the FP32 baseline. However, as in \fig{fig:method:fp16}, this reduction cannot be translated into real speedup without vectorized scatter/gather. 

NVIDIA GPUs group memory access requests into \textit{transactions}, whose largest size is 128 bytes. Considering the typical memory access pattern in scatter/gather, where a warp (32 threads) issues contiguous FP32 (4 bytes) memory access instructions simultaneously, the 128-byte transaction is fully utilized. However, when each thread in the warp issues an FP16 memory request, the memory transaction has only 64/128=50\% utilization, and the total number of memory transactions are essentially unchanged. As a result, we observe far smaller speedup (1.3$\times$) compared to the theoretical value (2$\times$) on scatter/gather if \emph{scalar} scatter/gather (\fig{fig:method:fp16}\textcolor{linkcolor}{a}) is performed. 

Contrarily, \emph{vectorized} scatter-gather~\fig{fig:method:fp16}\textcolor{linkcolor}{b} doubles the workload of each thread, making the total work of each warp still 128 bytes, equivalent to a full FP32 memory transaction. Meanwhile, the total number of memory transactions is \textit{halved} while the work for each memory transaction is \textit{unchanged}, and we observe \textbf{1.9$\times$} speedup over FP32 data movement on various GPU platforms. This closely aligns with the theoretical reduction in DRAM access.

Further quantizing the features to INT8 offers diminishing return, as the multi-way reduction in the scatter operation requires more than 8-bit for the final result. In this case, all scatter operations are still in 16 bits since CUDA requires aligned memory access. Thus, scattering (which takes 60\% of the data movement time) cannot not accelerated with the INT8 quantization, leading to limited overall speedup.

\subsubsection{Fused and Locality-Aware Memory Access}

Despite the limitation of aggressive feature quantization, it is still possible to achieve faster scatter/gather by exploiting locality. Intuitively, for a sparse convolution layer, the total amount of gather read and scatter write is $N_1 = |\mathcal{M}|(C_{\text{in}}+C_{\text{out}})$, where $\mathcal{M}$ is the map for this layer (defined in \sect{sect:background:mapping}), and $C_{\text{in}}$ and $C_{\text{out}}$ correspond to input and output channel numbers. However, the total feature size of this layer is $N_2 = N_{\text{in}}C_{\text{in}} + N_{\text{out}}C_{\text{out}}$. Empirically, the feature of each point is repetitively accessed for at least 4 times ($N_1 \geq 4N_2$). Based on this, we can ideally have \textbf{1.6$\times$} more DRAM access saving for scatter/gather (the amount of gather write and scatter read is also $N_1$ and cannot be saved).

As shown in \algo{alg:gather_mm_scatter} and \fig{fig:method:locality}\textcolor{linkcolor}{a}, the current implementation completely separates gather/scatter for different weights. When we perform \textit{gather} operation for $\bm{W}_{k+1}$, the GPU cache is filled with \textit{scatter} buffer features for $\bm{W}_k$ as long as the GPU cache size is much smaller than $N_1$ (typically $>$ 40MB, much larger than the 5.5MB L2 cache of NVIDIA RTX 2080 Ti). Intuitively, for gather operation on $\bm{W}_{k+1}$, we hope that the cache is filled with gather buffer features from $\bm{W}_k$. This suggests us to first fuse \textit{all} gather operations before \matmul, and fuse \textit{all} scatter operations afterwards. As such, the GPU cache will always hold data from the same type of buffer. 

Moreover, the memory access order matters. In the weight-stationary order (\fig{fig:method:locality}\textcolor{linkcolor}{a}), all map entries for weight $\bm{W}_k$ are \textit{unique}, so there is \textit{no} chance of feature reuse, and each gather/scatter leads to a cache miss. As in \fig{fig:method:locality}\textcolor{linkcolor}{b}, we instead take a \textit{locality-aware} memory access order. We gather the input features in the \textit{input-stationary} order and scatter the partial sums in the \textit{output-stationary} order.

Without loss of generality, we will focus on the implementation of input-stationary gather. We first maintain a neighbor set $\mathcal{N}_j$ for each input point $\bm{p}_j$: \ie, for the i\textsuperscript{th} map entry $(\bm{p}_j, \bm{q}_k, \bm{W}_n)$, we insert $(\bm{W}_n, i)$ into $\mathcal{N}_j$. Then, we iterate over every input point $\bm{p}_j$, fetch its feature vector $\bm{X}^\text{in}_j$ into the register, and write it to the corresponding DRAM location $\sum_{k=0}^{n-1} |\mathcal{M}[\bm{W}_k]| + i$ for each $(\bm{W}_n, i) \in \mathcal{N}_j$. Here, $\mathcal{M}[\bm{W}_k]$ is the map for weight $\bm{W}_k$. Note that each $\bm{X}^{\text{in}}_j$ is read from DRAM only once and held in the register. Hence, this algorithm achieves the optimal reuse for gather. Similar technique can be applied to scatter, where we read neighbors' partial sums for each output point from DRAM, perform reduction in the register, and write the result back only once. This optimization alone leads to \textbf{1.3-1.4$\times$} speedup in data movement on real-world point cloud datasets.

\subsection{Mapping Optimization}
\label{sect:method:sparsemapping}

\begin{figure}
    \centering
    \includegraphics[width=\linewidth]{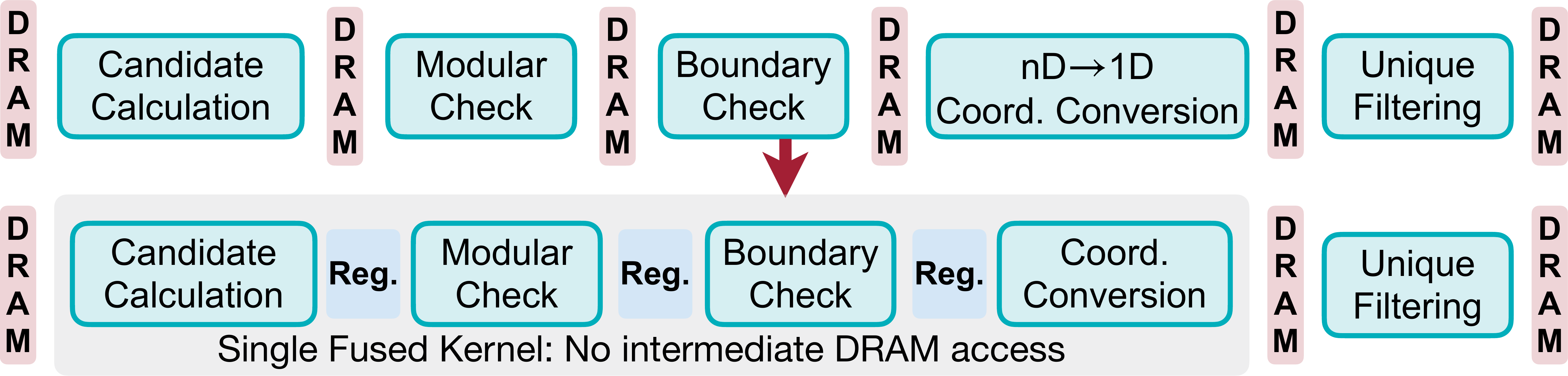}
    \caption{TorchSparse reduces \sparsemapping DRAM access and improves \sparsemapping latency via kernel fusion.}
    \label{fig:method:fused_kernels}
\end{figure}

\begin{figure*}[t]
    \centering
    \includegraphics[width=\linewidth]{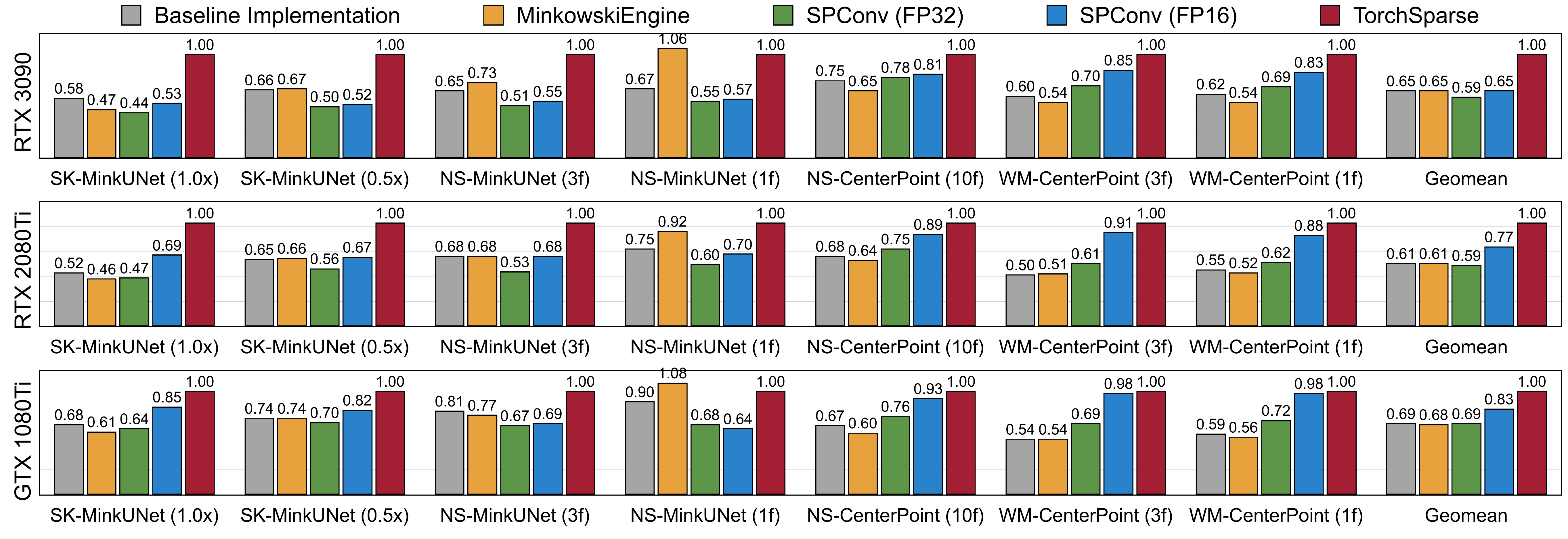}
    \caption{TorchSparse consistently outperforms state-of-the-art inference engines in both detection and segmentation benchmarks and achieves up to 1.5-1.6$\times$ geomean speedup, 2.3$\times$ single model speedup over \minkowskiengine and \spconv.}
    \label{fig:evaluation}
\end{figure*}

From \fig{fig:analysis:breakdown}, mapping operations in our baseline implementation take up a significant amount of time (15\%) in detectors on the Waymo~\cite{sun2020scalability} dataset. It is important to reduce the mapping overhead in sparse CNNs.

We first choose the map search strategy for each layer from [\texttt{grid}, \texttt{hashmap}] in a similar manner to the adaptive grouping. Here, \texttt{grid} corresponds to a naive collision-free grid-based hashmap: it takes larger memory space, but hashmap construction/query requires exactly one DRAM access per-entry, which is much smaller than conventional hashmaps. We then perform kernel fusion (\fig{fig:method:fused_kernels}) on output coordinates computation for downsampling. The downsample operation applies a sliding window around each point. It \circled{1} calculates candidate activated points with \texttt{broadcast\_add}, \circled{2} performs modular check, \circled{3} performs boundary check and generates a mask on whether each point is kept, \circled{4} converts the remaining candidate point coordinates to 1D values, and \circled{5} performs \texttt{unique} operation to keep final output coordinates (detailed in Appendix A). There are DRAM accesses between every two of the five stages, making downsampling kernels memory-bounded. We therefore fuse stages \circled{1} to \circled{4} into a single kernel and use registers to store intermediate results, which eliminates all intermediate DRAM write. For the fused kernel, we further perform control logic simplification, full loop unrolling and utilize the symmetry of submanifold maps. Overall, the mapping operations are accelerated by \textbf{4.6$\times$} on detection tasks with our optimizations.
\section{Evaluation}
\label{sect:evaluation}

\subsection{Setup}

TorchSparse is implemented in CUDA and provides easy-to-use PyTorch-like interfaces (described in \sect{sect:method:overview}). We build TorchSparse based on PyTorch 1.9.1 with CUDA 10.2/11.1 and cuDNN 7.6.5. Our system is evaluated against a baseline FP32 design without optimizations in \sect{sect:method} and the latest versions of two state-of-the-art sparse convolution libraries MinkowskiEngine v0.5.4~\cite{choy20194d} and SpConv v1.2.1~\cite{yan2018second} on three generations of NVIDIA GPUs: GTX 1080Ti, RTX 2080Ti and RTX 3090. Necessary changes are made to MinkowskiEngine to correctly support downsample operations in detectors and to SpConv to avoid OOM in large-scale scenes.

All systems are evaluated on seven top-performing sparse CNNs on large-scale datasets: MinkUNet~\cite{choy20194d} (0.5$\times$/1$\times$ width) on SemanticKITTI~\cite{behley2019semantickitti}, MinkUNet (1/3 frames) on nuScenes-LiDARSeg~\cite{caesar2020nuscenes}, CenterPoint~\cite{yin2021center} (10 frames) on nuScenes detection and CenterPoint (1/3 frames) on Waymo Open Dataset~\cite{sun2020scalability}. We report the normalized FPS for all systems (with TorchSparse to be 1).

\subsection{Evaluation Results}

\begin{figure}
\begin{subfigure}{0.48\linewidth}
    \centering
    \includegraphics[width=\linewidth]{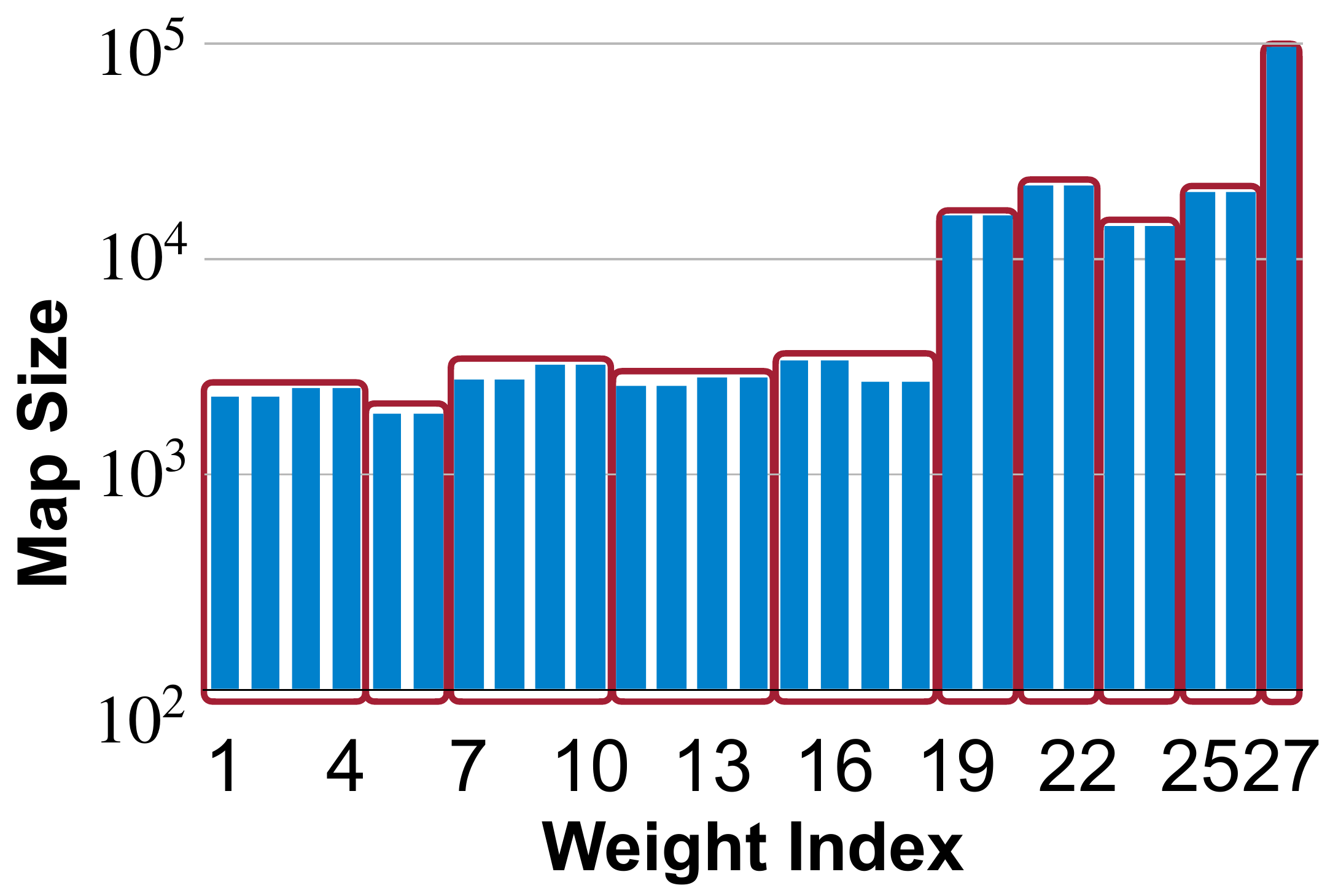}
    \caption{\Mappingpair size on SemanticKITTI}
    \label{fig:evaluation:nuscenes}
\end{subfigure}
\begin{subfigure}{0.48\linewidth}
    \centering
    \includegraphics[width=\linewidth]{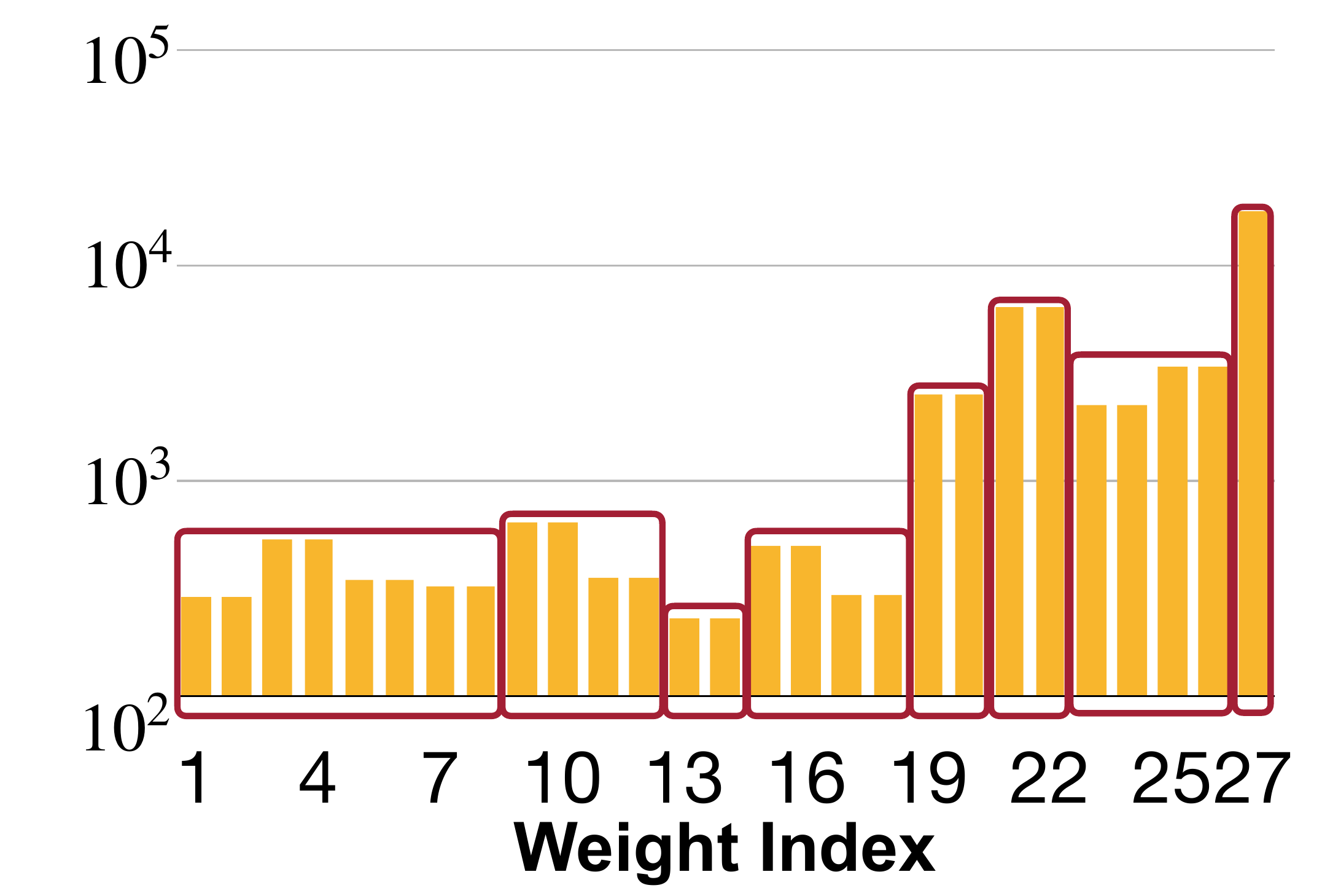}
    \caption{\Mappingpair size on nuScenes}
    \label{fig:evaluation:mmsize:semantickitti}
\end{subfigure}
\caption{Grouping strategy on different datasets. Maps on nuScenes are much smaller than on SemanticKITTI for MinkUNet. Thus, to fully utilize GPU, the grouping strategy is more aggressive on nuScenes (8 groups \vs 10 groups).}
\label{fig:ablation:mmsize}
\end{figure}

Our TorchSparse achieves the best performance compared with the baseline design, MinkowskiEngine and SpConv.

From \fig{fig:evaluation}, TorchSparse achieves up to 2.16$\times$ speedup on segmentation models and 1.6-2$\times$ speedup on detection models over MinkowskiEngine on RTX 3090. We achieve a smaller speedup for the 1-frame MinkUNet on nuScenes-LiDARSeg because MinkowskiEngine applies specialized optimizations to small models by using the \textit{fetch-on-demand} dataflow~\cite{lin2021pointacc} instead of the gather-matmul-scatter dataflow. 

TorchSparse also demonstrates a 1.2$\times$ faster inference speed compared with the FP16 version of SpConv for detectors on RTX3090 thanks to our fused and locality-aware access pattern and almost perfect speedup from vectorized data movement. Note that we report \textit{end-to-end} speedup in \fig{fig:evaluation}. However, 10\% of total total runtime in CenterPoint~\cite{yin2021center} is not related to point cloud computation (image convolution and non-maximum suppression, as in \fig{fig:analysis:breakdown}). Therefore, our speedup ratio on sparse convolution is 10\% more for CenterPoint. The performance gain over SpConv (FP16) is even larger on segmentation models on various hardware platforms thanks to the effectiveness of adaptively batched matrix multiplication, which will be discussed in \sect{sect:ablation:mm}. GPUs are usually more under-utilized for segmentation models as they usually have smaller workload compared with detectors, making it necessary to apply batching strategies to improve the device utilization.

TorchSparse achieves consistent speedup over other systems on GTX 1080Ti, which has \textit{no} FP16 tensor cores. Compared with the baseline design, our TorchSparse still achieves a 1.5$\times$ speedup, only 11\% less than the speedup we achieved on RTX 2080Ti with tensor cores. This validates that the native tensor-core speedup only constitutes a very minor proportion of our performance gain.

Our TorchSparse runs MinkUNet (1.0$\times$ on SemanticKITTI) at 36, 26 and 13 FPS on RTX 3090, RTX 2080Ti, GTX 1080Ti, respectively, all satisfying the real-time requirement ($\geq$ 10 FPS). For the 3-frame model on nuScenes-LiDARSeg, TorchSparse achieves 45, 40 and 25 FPS throughput on the three devices, at least 2$\times$ faster than the LiDAR frequency. Even for the heaviest 3-frame CenterPoint model on Waymo, our TorchSparse is still able to achieve the real-time inference on GTX 1080Ti. As such, our system paves the way for real-time LiDAR perception on self-driving cars.

\section{Ablation Study}
\label{sect:ablation}

\begin{table}[t]

\begin{subtable}{\linewidth}
\centering
\scalebox{0.8}{
\renewcommand{\arraystretch}{1.8}%
\addtolength{\tabcolsep}{6.3pt} 
\begin{tabular}{|c|c|c|c|}
\hline
\multicolumn{2}{|c|}{\multirow{2}{*}{
\makecell{Specialization \\ for Different \\ Datasets}}
} & \multicolumn{2}{c|}{Optimized for} \\ \cline{3-4} 
\multicolumn{2}{|c|}{}                      & SemanticKITTI & nuScenes \\ \hline
\multirow{2}{*}{\makecell{Execute \\ on}} & SemanticKITTI & \textbf{10.11}         & 10.87    \\ \cline{2-4} 
                            & nuScenes      & 5.30          & \textbf{4.67}     \\ \hline
\end{tabular}
}
\caption{Specialization for Datasets (MinkUNet, 2080Ti)}
\label{tab:ablation:mm:datasets}
\end{subtable}

\begin{subtable}{\linewidth}
\centering
\scalebox{0.8}{
\renewcommand{\arraystretch}{1.8}%
\begin{tabular}{|c|c|c|c|}
\hline
\multicolumn{2}{|c|}{\multirow{2}{*}{
\makecell{Specialization \\ for Different \\ Models}}
} & \multicolumn{2}{c|}{Optimized for} \\ \cline{3-4} 
\multicolumn{2}{|c|}{}  & MinkUNet (1.0$\times$) & MinkUNet (0.5$\times$) \\ \hline
\multirow{2}{*}{\makecell{Execute \\ on}} & MinkUNet (1.0$\times$) & \textbf{10.11}         & 10.70    \\ \cline{2-4} 
                            & MinkUNet (0.5$\times$)      & 5.37          & \textbf{4.72}     \\ \hline
\end{tabular}
}
\caption{Specialization for Model (SemanticKITTI, 2080Ti)}
\label{tab:ablation:mm:models}
\end{subtable}

\begin{subtable}{\linewidth}
\centering
\scalebox{0.8}{
\renewcommand{\arraystretch}{1.8}%
\addtolength{\tabcolsep}{8.7pt}
\begin{tabular}{|c|c|c|c|}
\hline
\multicolumn{2}{|c|}{\multirow{2}{*}{
\makecell{Specialization \\ for Different \\ Hardware}}
} & \multicolumn{2}{c|}{Optimized for} \\ \cline{3-4} 
\multicolumn{2}{|c|}{}  & RTX2080Ti & GTX1080Ti \\ \hline
\multirow{2}{*}{\makecell{Execute \\ on}} & RTX2080Ti & \textbf{4.67}         & 4.80    \\ \cline{2-4}
                            & GTX1080Ti      & 14.95          & \textbf{14.01}     \\ \hline
\end{tabular}
}
\caption{Specialization for Hardware (nuScenes, MinkUNet)}
\label{tab:ablation:mm:hardware}
\end{subtable}
\caption{Specializing adaptive batching strategies for different datasets, models and hardware platforms helps improve efficiency (TFLOP/s) by up to 13.5\%.}
\label{tab:ablation:mm_specialize}
\end{table}
\begin{table}[t]
\small\centering
\scalebox{0.85}{
\begin{tabular}{ccc}
\textbf{Grouping Method} & \textbf{\MM speedup (SK)} & \textbf{\MM Speedup (NS)} \\\midrule
Separate          & 8.1 TFLOP/s (1.00$\times$)    & 10.4 TFLOP/s (1.00$\times$)      \\
Symmetric         & 8.2 TFLOP/s (1.02$\times$)    & 14.6 TFLOP/s (1.39$\times$)      \\
Fixed             & 8.7 TFLOP/s (0.87$\times$)   & \textbf{21.1 TFLOP/s} (1.50$\times$)      \\
Adaptive          & \textbf{11.9 TFLOP/s (1.39$\times$)}    & 16.9 TFLOP/s (\textbf{1.54$\times$})    \\\bottomrule  
\end{tabular}
}
\caption{Ablation analysis on \matmul: adaptive batching consistently outperforms all other strategies in latency and brings about 1.4$\times$-1.5$\times$ speedup for \mm (SK=SemanticKITTI, NS=nuScenes). As we trade FLOPs for regularity, TFLOP/s and speedup are non-proportional. } 
\label{tab:ablation:mm}
\end{table}

\begin{table}[t]
\setlength{\tabcolsep}{2pt}
\centering
\small
\scalebox{0.83}{
\begin{tabular}{cccc|ccc}
\textbf{FP16} & \textbf{Vec.} & \textbf{Fused} & \textbf{Loc.-Aware} & \textbf{Speedup (G)} & \textbf{Speedup (S)} & \textbf{Speedup (SG)} \\\midrule
\textcolor{mydarkred}{\xmark}    & \textcolor{mydarkred}{\xmark}         & \textcolor{mydarkred}{\xmark}    & \textcolor{mydarkred}{\xmark}        & 1.00             & 1.00            & 1.00             \\
\textcolor{mydarkgreen}{\cmark}   & \textcolor{mydarkred}{\xmark}         & \textcolor{mydarkred}{\xmark}    & \textcolor{mydarkred}{\xmark}        &  1.17                &        1.48        &       1.32          \\
\textcolor{mydarkgreen}{\cmark}   & \textcolor{mydarkgreen}{\cmark}         & \textcolor{mydarkred}{\xmark}    & \textcolor{mydarkred}{\xmark}        &   1.91              &       1.95         &      1.93           \\
\textcolor{mydarkgreen}{\cmark}   & \textcolor{mydarkgreen}{\cmark}         & \textcolor{mydarkgreen}{\cmark}    & \textcolor{mydarkred}{\xmark}        &  1.91                &    2.12            &       2.02          \\
\textcolor{mydarkgreen}{\cmark}   & \textcolor{mydarkgreen}{\cmark}         & \textcolor{mydarkgreen}{\cmark}    & \textcolor{mydarkgreen}{\cmark}        &  \textbf{2.86}                &     \textbf{2.61}           &       \textbf{2.72}    \\\bottomrule     
\end{tabular}
}
\caption{Speedup breakdown of different optimizations to reduce data movement. Feature quantization, vectorized memory access, and fused and locality-aware access bring 1.3$\times$, 1.5$\times$ and 1.4$\times$ speedup, respectively. Here, \textit{G} and \textit{S} denote gather and scatter.}
\label{tab:ablation:scattergather}
\end{table}

\subsection{Matrix Multiplication Optimizations}
\label{sect:ablation:mm}

We first examine the performance of different grouping strategies on SemanticKITTI with MinkUNet (0.5$\times$) and on nuScenes with MinkUNet (3 frames). From \fig{fig:method:grouping}, our adaptive grouping strategy outperforms all handcrafted, fixed strategy and achieves 1.4-1.5$\times$ over no grouping baseline. \tbl{tab:ablation:mm} also suggests that manually-designed strategy cannot generalize to all datasets: fixed 3-batch grouping achieves large speedup (1.5$\times$) on nuScenes, but is 13\% slower than the separate computation baseline on SemanticKITTI. Note that although this strategy has the best device utilization (largest TFLOP/s) on nuScenes, it does not bring greater latency reduction than adaptive grouping due to much more extra computation, indicating the importance of $\epsilon$ in our adaptive grouping algorithm. We also show the effectiveness of grouping strategy specialization for different datasets, model and hardware in \tbl{tab:ablation:mm_specialize}. In \tbl{tab:ablation:mm:datasets}, we found that the same model (1-frame MinkUNet) on the same hardware platform benefits more from the dataset-specialized strategy. This is because map size distributions (which decides the workload of \matmul) significantly differ between SemanticKITTI and nuScenes, as shown in \fig{fig:ablation:mmsize}. The maps on nuScenes are much smaller than those on SemanticKITTI. As a result, if we directly transfer SemanticKITTI strategy to nuScenes, the groups will not be large enough to fully utilize hardware resources. On the other hand, if the nuScenes strategy is transferred to SemanticKITTI, the efficiency will be bottlenecked by computation overhead. We notice similar effect for model and hardware specialization in \tbl{tab:ablation:mm:models} and \tbl{tab:ablation:mm:hardware}, where specialized strategies always outperform the transferred ones.

\subsection{Data Movement Optimizations}
\label{sect:ablation:scattergather}

We then perform ablation analysis on MinkUNet~\cite{choy20194d} (1.0$\times$) on the SemanticKITTI dataset~\cite{behley2019semantickitti}. As in \tbl{tab:ablation:scattergather}, naively quantizing features to 16-bit will not provide significant speedup for scatter/gather: especially for gather, the speedup ratio is only 1.17$\times$, far less than the theoretical value (2$\times$). Instead, quantized and \textit{vectorized} scatter/gather improves the latency of scatter-gather by 1.93$\times$, which closely matches the DRAM access reduction and verifies our analysis in \sect{sect:method:scattergather} on memory transactions. We further observe that fusing gather/scatter itself will \textit{not} provide substantial speedup, as the \textit{weight-stationary} access pattern cannot provide good cache locality due to the uniqueness of maps for each weight. However, when combined with \textit{locality-aware} access, we achieve \textbf{2.86$\times$} speedup on gathering, \textbf{2.61$\times$} speedup on scattering and \textbf{2.72$\times$} overall speedup against FP32. This demonstrates the fact that all techniques in \sect{sect:method:scattergather} are crucial in improving the efficiency of data movement.

\subsection{Mapping Optimizations}
\label{sect:ablation:sparsemapping}

We finally present analysis on optimizing mapping operations in 3-frame CenterPoint~\cite{yin2021center} detector on Waymo~\cite{sun2020scalability}. Grid-based map search is 2.7$\times$ faster than a general hashmap-based solution thanks to its no-collision property, resulting in a 1.6$\times$ end-to-end speedup for mapping. Fusing four small kernels accelerates output construction by 2.1$\times$ and brings 1.5$\times$ further end-to-end mapping speedup. Finally, simplifying the control logic, loop unrolling and utilizing the symmetry of maps substantially accelerates map search by another 4$\times$ and pushes the final end-to-end mapping speedup to \textbf{4.6$\times$}.

\begin{figure}
    \centering
    \includegraphics[width=\linewidth]{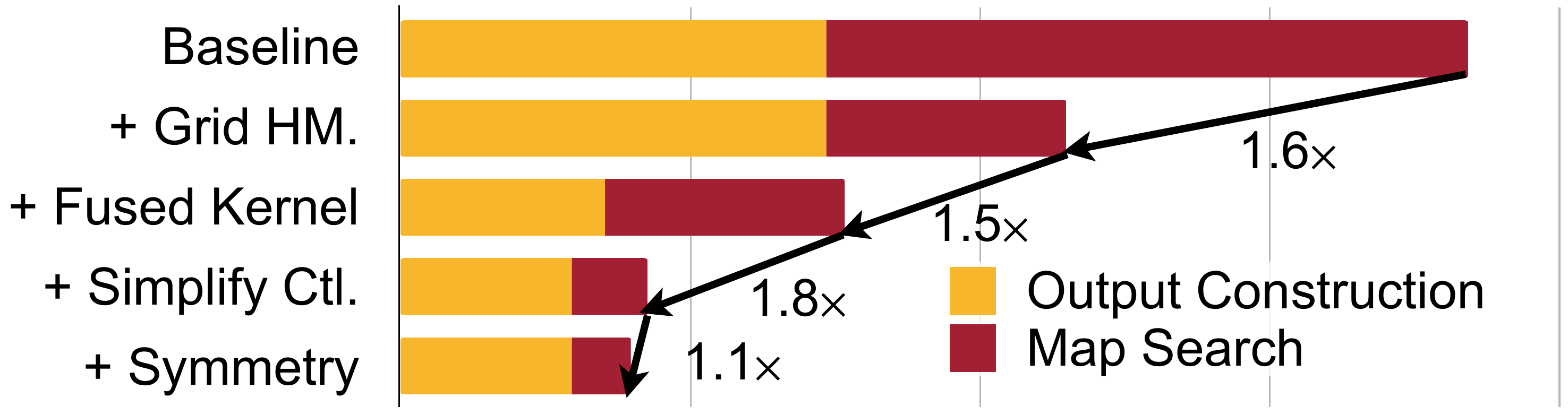}
    \caption{Speedup breakdown of mapping optimizations. Grid-based hashmap, fused kernel, simplified control logic and symmetry bring 1.6$\times$, 1.5$\times$, 1.8$\times$ and 1.1$\times$ measured speedup, respectively.}
    \label{fig:ablation:mapping}
\end{figure}

\section{Related Work}

\paragraph{Deep Learning on Point Clouds.}

Early methods~\cite{chang2015shapenet,qi2016volumetric,cicek20163d} first convert point clouds to the dense volumetric representation and apply dense CNNs to extract features. Another line of research ~\cite{qi2017pointnet,qi2017pointnet++,li2018pointcnn,wu2019pointconv,thomas2019kpconv,wang2018dgcnn} directly performs convolution on the $k$-nearest neighbor or spherical nearest neighbor of each point. Both streams of methods struggle to scale up to large scenes due to large or irregular memory footprint~\cite{liu2019pvcnn,liu2021pvnas}. Recent state-of-the-art deep learning methods on point cloud segmentation / detection~\cite{graham20183d,choy20194d,tang2020searching,shi2019pvrcnn,shi2021pvrcnn++,yin2021center} are usually based on sparse convolution, which is empirically proven to be able to scale up to large scenes and is the target for acceleration in this paper.

\paragraph{Point Cloud Inference Engines.}

Researchers have extensively developed efficient inference engines for sparse convolution inference. SpConv~\cite{yan2018second} proposes grid-based map search and the gather-matmul-scatter dataflow. SparseConvNet~\cite{graham20183d} proposes hashmap-based map search and is later significantly improved (in latency) by MinkowskiEngine, which also introduces a new \textit{fetch-on-demand} dataflow that excels at small workloads and allows generalized sparse convolution on $>$3D point clouds and on arbitrary coordinates.
\section{Conclusion}
\label{sect:conclusion}

We present TorchSparse, an open-source inference engine for efficient point cloud neural networks. Guided by two general principles: trade computation for regularity and reduce memory footprint, we optimize \matmul, data movement and \sparsemapping operations in sparse convolutions, achieving up to 1.5$\times$, 2.7$\times$ and 4.6$\times$ speedup on these three components, and up to 1.5-1.6$\times$ end-to-end speedup over previous state-of-the-art point cloud inference engines on both segmentation and detection tasks. We hope that our in-depth analysis on the efficiency bottlenecks and optimization recipes for sparse convolution can inspire future research on point cloud inference engine design.

\section*{Acknowledgments}

We would like to thank Hanrui Wang and Ligeng Zhu for their feedback on the artifact evaluation. This research was supported by NSF CAREER Award \#1943349, Ford and Hyundai. Zhijian Liu and Yujun Lin were partially supported by the Qualcomm Innovation Fellowship.

\bibliography{reference}

\begin{thebibliography}{25}
\providecommand{\natexlab}[1]{#1}
\providecommand{\url}[1]{\texttt{#1}}
\expandafter\ifx\csname urlstyle\endcsname\relax
  \providecommand{\doi}[1]{doi: #1}\else
  \providecommand{\doi}{doi: \begingroup \urlstyle{rm}\Url}\fi

\bibitem[Behley et~al.(2019)Behley, Garbade, Milioto, Quenzel, Behnke,
  Stachniss, and Gall]{behley2019semantickitti}
Behley, J., Garbade, M., Milioto, A., Quenzel, J., Behnke, S., Stachniss, C.,
  and Gall, J.
\newblock {SemanticKITTI: A Dataset for Semantic Scene Understanding of LiDAR
  Sequences}.
\newblock In \emph{IEEE/CVF International Conference on Computer Vision
  (ICCV)}, 2019.

\bibitem[Caesar et~al.(2020)Caesar, Bankiti, Lang, Vora, Liong, Xu, Krishnan,
  Pan, Baldan, and Beijbom]{caesar2020nuscenes}
Caesar, H., Bankiti, V., Lang, A.~H., Vora, S., Liong, V.~E., Xu, Q., Krishnan,
  A., Pan, Y., Baldan, G., and Beijbom, O.
\newblock {nuScenes: A Multimodal Dataset for Autonomous Driving}.
\newblock In \emph{IEEE/CVF Conference on Computer Vision and Pattern
  Recognition (CVPR)}, 2020.

\bibitem[Chang et~al.(2015)Chang, Funkhouser, Guibas, Hanrahan, Huang, Li,
  Savarese, Savva, Song, Su, Xiao, Yi, and Yu]{chang2015shapenet}
Chang, A.~X., Funkhouser, T., Guibas, L., Hanrahan, P., Huang, Q., Li, Z.,
  Savarese, S., Savva, M., Song, S., Su, H., Xiao, J., Yi, L., and Yu, F.
\newblock {ShapeNet: An Information-Rich 3D Model Repository}.
\newblock \emph{arXiv}, 2015.

\bibitem[Choy et~al.(2019)Choy, Gwak, and Savarese]{choy20194d}
Choy, C., Gwak, J., and Savarese, S.
\newblock {4D Spatio-Temporal ConvNets: Minkowski Convolutional Neural
  Networks}.
\newblock In \emph{IEEE/CVF Conference on Computer Vision and Pattern
  Recognition (CVPR)}, 2019.

\bibitem[Cicek et~al.(2016)Cicek, Abdulkadir, Lienkamp, Brox, and
  Ronneberger]{cicek20163d}
Cicek, O., Abdulkadir, A., Lienkamp, S.~S., Brox, T., and Ronneberger, O.
\newblock {3D U-Net: Learning Dense Volumetric Segmentation from Sparse
  Annotation}.
\newblock In \emph{Proc. Medical Image Computing and Computer Assisted
  Intervention (MICCAI)}, 2016.

\bibitem[Ge et~al.(2021)Ge, Ding, Hu, Shao, Huang, Li, and Liu]{ge2021afdet}
Ge, R., Ding, Z., Hu, Y., Shao, W., Huang, L., Li, K., and Liu, Q.
\newblock {1\textsuperscript{st} Place Solutions to the Real-time 3D Detection
  and the Most Efficient Model of the Waymo Open Dataset Challenge 2021}.
\newblock In \emph{IEEE/CVF Conference on Computer Vision and Pattern
  Recognition Workshops (CVPRW)}, 2021.

\bibitem[Graham et~al.(2018)Graham, Engelcke, and van~der Maaten]{graham20183d}
Graham, B., Engelcke, M., and van~der Maaten, L.
\newblock {3D Semantic Segmentation With Submanifold Sparse Convolutional
  Networks}.
\newblock In \emph{IEEE/CVF Conference on Computer Vision and Pattern
  Recognition (CVPR)}, 2018.

\bibitem[Hu et~al.(2020)Hu, Ye, Wang, Yu, Zheng, Li, Zhang, Zhang, and
  Wang]{hu2020featgraph}
Hu, Y., Ye, Z., Wang, M., Yu, J., Zheng, D., Li, M., Zhang, Z., Zhang, Z., and
  Wang, Y.
\newblock {FeatGraph: A Flexible and Efficient Backend for Graph Neural Network
  Systems}.
\newblock In \emph{International Conference for High Performance Computing,
  Networking, Storage and Analysis (SC)}, 2020.

\bibitem[Li et~al.(2018)Li, Bu, Sun, Wu, Di, and Chen]{li2018pointcnn}
Li, Y., Bu, R., Sun, M., Wu, W., Di, X., and Chen, B.
\newblock {PointCNN: Convolution on $\mathcal{X}$-Transformed Points}.
\newblock In \emph{Advances in Neural Information Processing Systems
  (NeurIPS)}, 2018.

\bibitem[Lin et~al.(2021)Lin, Zhang, Tang, Wang, and Han]{lin2021pointacc}
Lin, Y., Zhang, Z., Tang, H., Wang, H., and Han, S.
\newblock {PointAcc: Efficient Point Cloud Accelerator}.
\newblock In \emph{54th Annual IEEE/ACM International Symposium on
  Microarchitecture (MICRO)}, 2021.

\bibitem[Liu et~al.(2019)Liu, Tang, Lin, and Han]{liu2019pvcnn}
Liu, Z., Tang, H., Lin, Y., and Han, S.
\newblock {Point-Voxel CNN for Efficient 3D Deep Learning}.
\newblock In \emph{Advances in Neural Information Processing Systems
  (NeurIPS)}, 2019.

\bibitem[Liu et~al.(2021)Liu, Tang, Zhao, Shao, and Han]{liu2021pvnas}
Liu, Z., Tang, H., Zhao, S., Shao, K., and Han, S.
\newblock {PVNAS: 3D Neural Architecture Search with Point-Voxel Convolution}.
\newblock \emph{IEEE Transactions on Pattern Analysis and Machine Intelligence
  (TPAMI)}, 2021.

\bibitem[Qi et~al.(2016)Qi, Su, Niessner, Dai, Yan, and
  Guibas]{qi2016volumetric}
Qi, C.~R., Su, H., Niessner, M., Dai, A., Yan, M., and Guibas, L.~J.
\newblock {Volumetric and Multi-View CNNs for Object Classification on 3D
  Data}.
\newblock In \emph{IEEE/CVF Conference on Computer Vision and Pattern
  Recognition (CVPR)}, 2016.

\bibitem[Qi et~al.(2017{\natexlab{a}})Qi, Su, Mo, and Guibas]{qi2017pointnet}
Qi, C.~R., Su, H., Mo, K., and Guibas, L.~J.
\newblock {PointNet: Deep Learning on Point Sets for 3D Classification and
  Segmentation}.
\newblock In \emph{IEEE/CVF Conference on Computer Vision and Pattern
  Recognition (CVPR)}, 2017{\natexlab{a}}.

\bibitem[Qi et~al.(2017{\natexlab{b}})Qi, Yi, Su, and Guibas]{qi2017pointnet++}
Qi, C.~R., Yi, L., Su, H., and Guibas, L.~J.
\newblock {PointNet++: Deep Hierarchical Feature Learning on Point Sets in a
  Metric Space}.
\newblock In \emph{Advances in Neural Information Processing Systems
  (NeurIPS)}, 2017{\natexlab{b}}.

\bibitem[Shi et~al.(2020)Shi, Guo, Jiang, Wang, Shi, Wang, and
  Li]{shi2019pvrcnn}
Shi, S., Guo, C., Jiang, L., Wang, Z., Shi, J., Wang, X., and Li, H.
\newblock {PV-RCNN: Point-Voxel Feature Set Abstraction for 3D Object
  Detection}.
\newblock In \emph{IEEE/CVF Conference on Computer Vision and Pattern
  Recognition (CVPR)}, 2020.

\bibitem[Shi et~al.(2021)Shi, Jiang, Deng, Wang, Guo, Shi, Wang, and
  Li]{shi2021pvrcnn++}
Shi, S., Jiang, L., Deng, J., Wang, Z., Guo, C., Shi, J., Wang, X., and Li, H.
\newblock {PV-RCNN++: Point-Voxel Feature Set Abstraction With Local Vector
  Representation for 3D Object Detection}.
\newblock \emph{arXiv preprint arXiv:2102.00463}, 2021.

\bibitem[Sun et~al.(2020)Sun, Kretzschmar, Dotiwalla, Chouard, Patnaik, Tsui,
  Guo, Zhou, Chai, Caine, Vasudevan, Han, Ngiam, Zhao, Timofeev, Ettinger,
  Krivokon, Gao, Joshi, Zhang, Shlens, Chen, and Anguelov]{sun2020scalability}
Sun, P., Kretzschmar, H., Dotiwalla, X., Chouard, A., Patnaik, V., Tsui, P.,
  Guo, J., Zhou, Y., Chai, Y., Caine, B., Vasudevan, V., Han, W., Ngiam, J.,
  Zhao, H., Timofeev, A., Ettinger, S., Krivokon, M., Gao, A., Joshi, A.,
  Zhang, Y., Shlens, J., Chen, Z., and Anguelov, D.
\newblock {Scalability in Perception for Autonomous Driving: Waymo Open
  Dataset}.
\newblock In \emph{IEEE/CVF Conference on Computer Vision and Pattern
  Recognition (CVPR)}, 2020.

\bibitem[Tang et~al.(2020)Tang, Liu, Zhao, Lin, Lin, Wang, and
  Han]{tang2020searching}
Tang, H., Liu, Z., Zhao, S., Lin, Y., Lin, J., Wang, H., and Han, S.
\newblock {Searching Efficient 3D Architectures with Sparse Point-Voxel
  Convolution}.
\newblock In \emph{European Conference on Computer Vision (ECCV)}, 2020.

\bibitem[Thomas et~al.(2019)Thomas, Qi, Deschaud, Marcotegui, Goulette, and
  Guibas]{thomas2019kpconv}
Thomas, H., Qi, C.~R., Deschaud, J.-E., Marcotegui, B., Goulette, F., and
  Guibas, L.~J.
\newblock {KPConv: Flexible and Deformable Convolution for Point Clouds}.
\newblock In \emph{IEEE/CVF International Conference on Computer Vision
  (ICCV)}, 2019.

\bibitem[Wang et~al.(2019{\natexlab{a}})Wang, Zheng, Ye, Gan, Li, Song, Zhou,
  Ma, Yu, Gai, Xiao, He, Karypis, Lin, and Zhang]{wang2019dgl}
Wang, M., Zheng, D., Ye, Z., Gan, Q., Li, M., Song, X., Zhou, J., Ma, C., Yu,
  L., Gai, Y., Xiao, T., He, T., Karypis, G., Lin, J., and Zhang, Z.
\newblock {Deep Graph Library: A Graph-Centric, Highly-Performant Package for
  Graph Neural Networks}.
\newblock \emph{arXiv preprint arXiv:1909.01315}, 2019{\natexlab{a}}.

\bibitem[Wang et~al.(2019{\natexlab{b}})Wang, Sun, Liu, Sarma, Bronstein, and
  Solomon]{wang2018dgcnn}
Wang, Y., Sun, Y., Liu, Z., Sarma, S.~E., Bronstein, M.~M., and Solomon, J.~M.
\newblock {Dynamic Graph CNN for Learning on Point Clouds}.
\newblock In \emph{ACM SIGGRAPH}, 2019{\natexlab{b}}.

\bibitem[Wu et~al.(2019)Wu, Qi, and Fuxin]{wu2019pointconv}
Wu, W., Qi, Z., and Fuxin, L.
\newblock {PointConv: Deep Convolutional Networks on 3D Point Clouds}.
\newblock In \emph{IEEE/CVF Conference on Computer Vision and Pattern
  Recognition (CVPR)}, 2019.

\bibitem[Yan et~al.(2018)Yan, Mao, and Li]{yan2018second}
Yan, Y., Mao, Y., and Li, B.
\newblock {SECOND: Sparsely Embedded Convolutional Detection}.
\newblock \emph{Sensors}, 2018.

\bibitem[Yin et~al.(2021)Yin, Zhou, and Kr\"ahenb\"uhl]{yin2021center}
Yin, T., Zhou, X., and Kr\"ahenb\"uhl, P.
\newblock {Center-based 3D Object Detection and Tracking}.
\newblock In \emph{IEEE/CVF Conference on Computer Vision and Pattern
  Recognition (CVPR)}, 2021.

\end{thebibliography}
\bibliographystyle{mlsys2022}

\clearpage
\appendix
\section{Output Coordinates Calculation}

Here, we illustrate the output coordinates calculation algorithm for $s > 1$ in sparse convolution. We apply a sliding window on each input point and check whether each candidate output point within the window passes modular and boundary check. If both checks are passed, we add the candidate output point to $\bm{P}^{\text{out}}$. We finally filter out duplicate coordinates in $\bm{P}^{\text{out}}$.

\begin{algorithm}
\caption{Output Coordinates Calculation}\label{alg:output_coords_calc}
\textbf{Input:} kernel size $K$, stride $s$, input coordinates $\bm{P}^{\text{in}}$, \\
\hspace*{2.8em} output coordinates boundary $\bm{b}$\\
\textbf{Output:} output coordinates $\bm{P}^{\text{out}}$
\begin{algorithmic}
\IF {$s = 1$} 
  \STATE $\bm{P}^{\text{out}}\gets \bm{P}^{\text{in}}$
\ELSE
  \STATE $\bm{P}^{\text{out}}\gets \emptyset$
  \FOR {$\bm{p}$ \textbf{in} $\bm{P}^{\text{in}}$}
    \STATE \small{\textcolor{blue}{\# Traverse the neighbors of an input point.}}
    \FOR{$\bm{\delta}$ \textbf{in} $\OffsetSet{D}{K}$} 
    \STATE \small{\textcolor{blue}{\# Calculate the candidate coordinates.}}
    \STATE $\bm{u}\gets \bm{p} - \bm{\delta}$
    \STATE \small{\textcolor{blue}{\# Add output if it passes modular and boundary check.}}
    \IF {$\bm{u}\;\%\;s == 0$ \textbf{and} $\bm{u} < s\cdot\bm{b}$}
      \STATE $\bm{P}^{\text{out}}\gets \bm{P}^{\text{out}} \cup \{\bm{u} / s\} $  
    \ENDIF
    \ENDFOR
  \ENDFOR
  \STATE \small{\textcolor{blue}{\# Filter out duplicate coordinates.}}
  \STATE $\bm{P}^{\text{out}}\gets \texttt{Unique}(\bm{P}^{\text{out}})$
\ENDIF
\end{algorithmic}
\end{algorithm}

\section{Adaptive Grouping Algorithm}

\begin{figure*}[t]
    \centering
    \includegraphics[width=\linewidth]{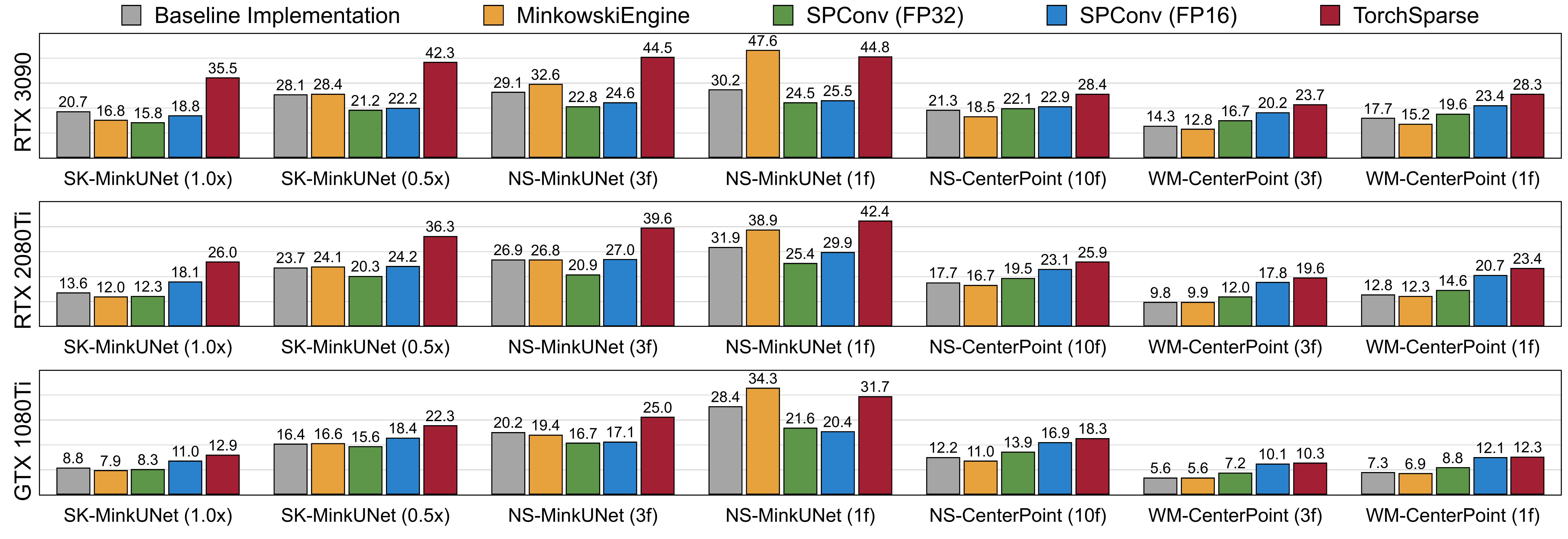}
    \caption{TorchSparse evaluation results in \textit{absolute} values.}
    \label{fig:appendix:evaluation}
\end{figure*}

Here, we provide detailed illustration for the adaptive grouping algorithm. The algorithm is divided into two parts: grouped matrix multiplication (\algo{alg:group_matmul}) and adaptive group search (\algo{alg:adap_group_search}).

\subsection{Group Matrix Multiplication}
\label{sect:method:group_matmul}

We describe the process of applying the adaptive grouping strategies for each layer in \algo{alg:group_matmul}, which is performed via two steps. First, we maintain two pointers to track the start and the end of the current group. Once $1 - n_{\min}/ n_{\max}$ updated by the end pointer exceeds the tolerance of redundant computation $\epsilon$, we return the working group to the groups list and move pointers to start a new group. Second, for each group, we determine if batched matmul is performed on it based on the value of $S$.

\begin{algorithm}
\caption{Grouped \MatMul}\label{alg:group_matmul}
\textbf{Input:} input features $\bm{X}^{\text{in}}$, weights $\bm{W}$, \mappingpairs $\mathcal{M}$,\\
\hspace*{2.8em} redundant computation tolerance $\epsilon$,\\
\hspace*{2.8em} $\texttt{mm}$/$\texttt{bmm}$ threshold $S$\\
\textbf{Output:} output features $\bm{X}^{\text{out}}$
\begin{algorithmic}
\STATE $\bm{G} \gets \emptyset$
\STATE $i \gets 0$
\STATE \small{\textcolor{blue}{\# Traverse each weight index with unique number of inputs.}}
\WHILE{$i < \texttt{range}(\lfloor\OffsetSet{D}{K}.\texttt{size} / 2\rfloor)$}
    \STATE $n_{\min} \gets 0; \ n_{\max} \gets \texttt{len}(\mathcal{M}[\bm{W}_i])$
    \STATE \small{\textcolor{blue}{\# Always push the first index to the current group.}}
    \STATE $\bm{g} \gets \{i\}; \ i \gets i+1$
    
    \FOR{$j$ \textbf{in} $\texttt{range}(i, \lfloor\OffsetSet{D}{K}.\texttt{size} / 2\rfloor)$}
        \STATE $n \gets \texttt{len}(\mathcal{M}[\bm{W}_j])$
        \STATE $n_{\min} \gets \min\{n, n_{\min}\}, \ n_{\max} \gets \max\{n, n_{\max}\}$
        \STATE \small{\textcolor{blue}{\# Push the index to the group if the ratio no larger than $\epsilon$.}}
        \IF{$1 - n_{\min} / n_{\max}  \leq  \epsilon$}
            \STATE $\bm{g} \gets \bm{g} \cup \{j\}$
        \ELSE
            \STATE \small{\textcolor{blue}{\# Otherwise return and start a new group.}}
            \STATE \textbf{break}
        \ENDIF
    \ENDFOR
    \STATE \small{\textcolor{blue}{\# Push the returned group to the groups list.}}
    \STATE $\bm{G} \gets \bm{G} \cup \{\bm{g}\}$
\ENDWHILE

\FOR{$\bm{g}$ \textbf{in} $\bm{G}$}
    \STATE $n_{\max} \gets \max\{\texttt{len}(\mathcal{M}[\bm{W}_i]) \ \text{for} \ i \  \text{in} \ \bm{g}\}$
    \STATE \small{\textcolor{blue}{\# Pad inputs and apply \texttt{bmm} when workload smaller than $S$.}}
    \IF{$n_{\max} < S$}
        \STATE gather input feature matrices $\bm{F}_{i}$ with $\bm{X}^{\text{in}}$ and $\mathcal{M}[\bm{W}_i]$ for $i \in \bm{g}$; pad zeros to each $\bm{F}_{i}$ to become length $n_{max}$; perform batched matrix multiplication between $\bm{F}_{i \in \bm{g}}$ and $\bm{W}_{i \in \bm{g}}$ and then scatter results to corresponding $\bm{X}^{\text{out}}$
    \ELSE
        \STATE \small{\textcolor{blue}{\# Otherwise apply \texttt{mm}.}}
        \STATE perform \algo{alg:gather_mm_scatter} in main paper with $\bm{X}^{\text{in}}_{i}$ for $i \in \bm{g}$ to get $\bm{X}^{\text{out}}$
    \ENDIF
\ENDFOR
\end{algorithmic}
\end{algorithm}

\subsection{Adaptive Strategy Search}
\label{sect:method:adap_group_search}

For each layer, we search for a specific configuration to conduct adaptive grouping (i.e. auto-tune $\epsilon$ and $S$). The tuning algorithm is shown in \algo{alg:adap_group_search}, where we enumerate $\epsilon, S$ in a predefined search space (usually $<$ 1000 configurations), use \algo{alg:group_matmul} to perform the matrix multiplication for the target layer, and select the $\epsilon, S$ pair which leads to the smallest average latency.

\begin{algorithm}
\caption{Adaptive Group Search}\label{alg:adap_group_search}
\textbf{Input:} sampled inputs subset $\bm{D}$, weights $\bm{W}$, \mappingpairs $\mathcal{M}$,\\
\hspace*{2.8em} redundant computation tolerance search space $\mathcal{S}_{a}$,\\ 
\hspace*{2.8em} $\texttt{mm}$/$\texttt{bmm}$ threshold search space $\mathcal{S}_{b}$\\
\textbf{Output:}  selected redundant computation tolerance $\epsilon^{*}$, \\
\hspace*{2.8em} $\texttt{mm}$/$\texttt{bmm}$ threshold $S^{*}$
\begin{algorithmic}
\STATE $f \gets$ {cost function to compute elapsed time on hardware}
\STATE $c_{\min} \gets 0$
\FOR{$\epsilon$ \textbf{in} $\mathcal{S}_{a}$}
    \FOR{$S$ \textbf{in} $\mathcal{S}_{b}$}
        \STATE $c \gets 0$
        \FOR{$\bm{X}^{\text{in}}$ \textbf{in} $\bm{D}$}
            \STATE $c \gets c + f(\text{run \algo{alg:group_matmul} with }\bm{X}^{\text{in}}, \bm{W}, \mathcal{M}, \epsilon, S)$
        \ENDFOR
        \STATE \small{\textcolor{blue}{\# Update selected config to $\{\epsilon, S\}$ for smaller latency.}}
        \IF{$c_{\min} = 0 \ \algorithmicor \ c < c_{\min}$}
            \STATE $c_{\min} \gets c$
            \STATE $\epsilon^{*} \gets \epsilon, S^{*} \gets S$
        \ENDIF
    \ENDFOR
\ENDFOR
\end{algorithmic}
\end{algorithm}

\section{Results Detail}

We show TorchSparse evaluation results in absolute FPS in \fig{fig:appendix:evaluation}. TorchSparse is able to run all models in real-time ($>$ 10 FPS) on all hardware platforms.

\end{document}